\documentclass{article}

\usepackage{arxiv}

\usepackage[utf8]{inputenc} % allow utf-8 input
\usepackage[T1]{fontenc}    % use 8-bit T1 fonts
\usepackage{hyperref}       % hyperlinks
\usepackage{url}            % simple URL typesetting
\usepackage{booktabs}       % professional-quality tables
\usepackage{amsfonts}       % blackboard math symbols
\usepackage{nicefrac}       % compact symbols for 1/2, etc.
\usepackage{microtype}      % microtypography
\usepackage{lipsum}		% Can be removed after putting your text content
\usepackage{graphicx}
\usepackage{natbib}
\usepackage{doi}
\usepackage{amsmath}
\usepackage{algorithm}
\usepackage{algpseudocode}

\title{TabularFM: An Open Framework For Tabular Foundational Models}

\date{} 					% Or removing it

\author{ Quan M. Tran\thanks{Equal contribution} \\
	Vietnam National University\\
        Ho Chi Minh City, Vietnam\\
	\And
	Suong N. Hoang\footnotemark[1] \\
	Vietnam National University\\ 
        Ho Chi Minh City, Vietnam \\
	\AND
	Lam M. Nguyen \\
	IBM Research \\
	Yorktown Height, US \\
	\And
	Dzung Phan \\
	IBM Research \\
	Yorktown Height, US \\
	\And
	Hoang Thanh Lam \\
	IBM Research \\
	Dublin, Ireland \\
}

\renewcommand{\shorttitle}{TabularFM: An Open Framework For Tabular Foundational Models}

\begin{document}
\maketitle

\begin{abstract}
Foundational models (FMs), pretrained on extensive datasets using self-supervised techniques, are capable of learning generalized patterns from large amounts of data. This reduces the need for extensive labeled datasets for each new task, saving both time and resources by leveraging the broad knowledge base established during pretraining. Most research on FMs has primarily focused on unstructured data, such as text and images, or semi-structured data, like time-series. However, there has been limited attention to structured data, such as tabular data, which, despite its prevalence, remains under-studied due to a lack of clean datasets and insufficient research on the transferability of FMs for various tabular data tasks. In response to this gap, we introduce a framework called TabularFM\footnote{\url{https://tabularfm.github.io/}}, which incorporates state-of-the-art methods for developing FMs specifically for tabular data. This includes variations of neural architectures such as GANs, VAEs, and Transformers. We have curated a million of tabular datasets and released cleaned versions to facilitate the development of tabular FMs. We pretrained FMs on this curated data, benchmarked various learning methods on these datasets, and released the pretrained models along with leaderboards for future comparative studies. Our fully open-sourced system provides a comprehensive analysis of the transferability of tabular FMs. By releasing these datasets, pretrained models, and leaderboards, we aim to enhance the validity and usability of tabular FMs in the near future.
\end{abstract}

% keywords can be removed
% \keywords{First keyword \and Second keyword \and More}

\section{Introduction}
Foundational models  (FMs) undergo training on extensive datasets through self-supervised learning methods. These models learns crucial patterns and structures within the data autonomously, without human guidance. Following this, they are fine-tuned on downstream tasks utilizing smaller datasets with labeled data. Research has shown that if foundational models are pre-trained on large datasets resembling the downstream datasets, the learned patterns from pre-training can be leveraged for the downstream tasks, resulting in a substantial performance boost for the subsequent models \cite{he2016deep, waswani2017attention, lin2023evolutionary}. 

Foundational models have been proposed to handle unstructured data across various domains, including vision \cite{he2016deep}, text \cite{waswani2017attention}, and biomedical data like proteins \cite{lin2023evolutionary} or small molecules \cite{10.1038/s42256-022-00580-7}. However, when it comes to structured data such as tables, the task of constructing foundational models remains largely unexplored. Several challenges arise in building foundational models for tabular data. First, the transferability of learning methods across tabular datasets remains an open question. Unlike texts or images, where neural architectures like CNNs \cite{lecun2015deep} or Transformers \cite{waswani2017attention} were introduced to learn transferable patterns, tabular data presents distinct challenges. These include variations in categorical value encoding and numerical value scales, resulting in higher noise levels. Second, despite the existence of public tabular datasets, they tend to be small and noisy. For instance, some contain a mixture of text data or temporal information, which diverges from structured data. Moreover, the sizes of these available datasets are considerably smaller compared to the texts used in creating large language models. As this is an emerging area of research, there is a lack of standard benchmarks and leaderboards for evaluating tabular foundational models. In a recent position paper \citep{vanbreugel2024tabular}, the authors noticed that tabular foundation models remain largely unexplored compared to the abundance of text and image foundation models. For instance, the number of accepted papers on tabular foundation models at recent major machine learning conferences is less than three percent of those focused on text foundation models. In this work, we present TabularFM, a comprehensive framework for creating and benchmarking tabular foundational models. Our contributions can be summarised as follows:

\begin{itemize}
\item Datasets for training and benchmark tabular FMs: clean datasets comprising a total of 2,693 tables, meticulously curated from 1 million GitTables \cite{hulsebos2023gittables} and 43,514 tables crawled from Kaggle\footnote{www.kaggle.com}. We released pretrained FM models on the curated datasets, benchmark datasets, and associated leaderboards for evaluating tabular FMs.
\item Additionally, we provide an open-source framework for creating and benchmarking tabular FMs. This framework encompasses state-of-the-art self-supervised generative learning methods, data transformation techniques, and evaluation tools.

\item An analysis of the transferability of tabular foundation models highlights several key findings. In Subsection \ref{sec:Model Transferability Comparison}, we show that pretrained models using CTGAN and TVAE transfer effectively across various data splits, outperforming models trained from scratch. However, adding extra column metadata does not improve transferability. Transformers pretrained on textual data perform well on benchmarks, but finetuning them on tabular data often reduces performance, raising questions about the need for larger tabular datasets. Our analysis in Figure \ref{fig:correlation} indicates that certain general knowledge, like the correlation between \texttt{hospital beds} and \texttt{urban population} or the link between \texttt{weight loss} and \texttt{polydipsia} via \texttt{polyuria}, transfers well through pretraining.
\end{itemize}

\section{Data Creation}
In this section, we briefly discuss the process of data creation, which includes data acquisition and data cleaning. Since building foundational models requires clean data, this step is crucial to ensure that only high-quality data is retained for training.
\subsection{Kaggle Dataset}
\begin{figure}[h]
    \centering
    \includegraphics[width=1.0\textwidth]{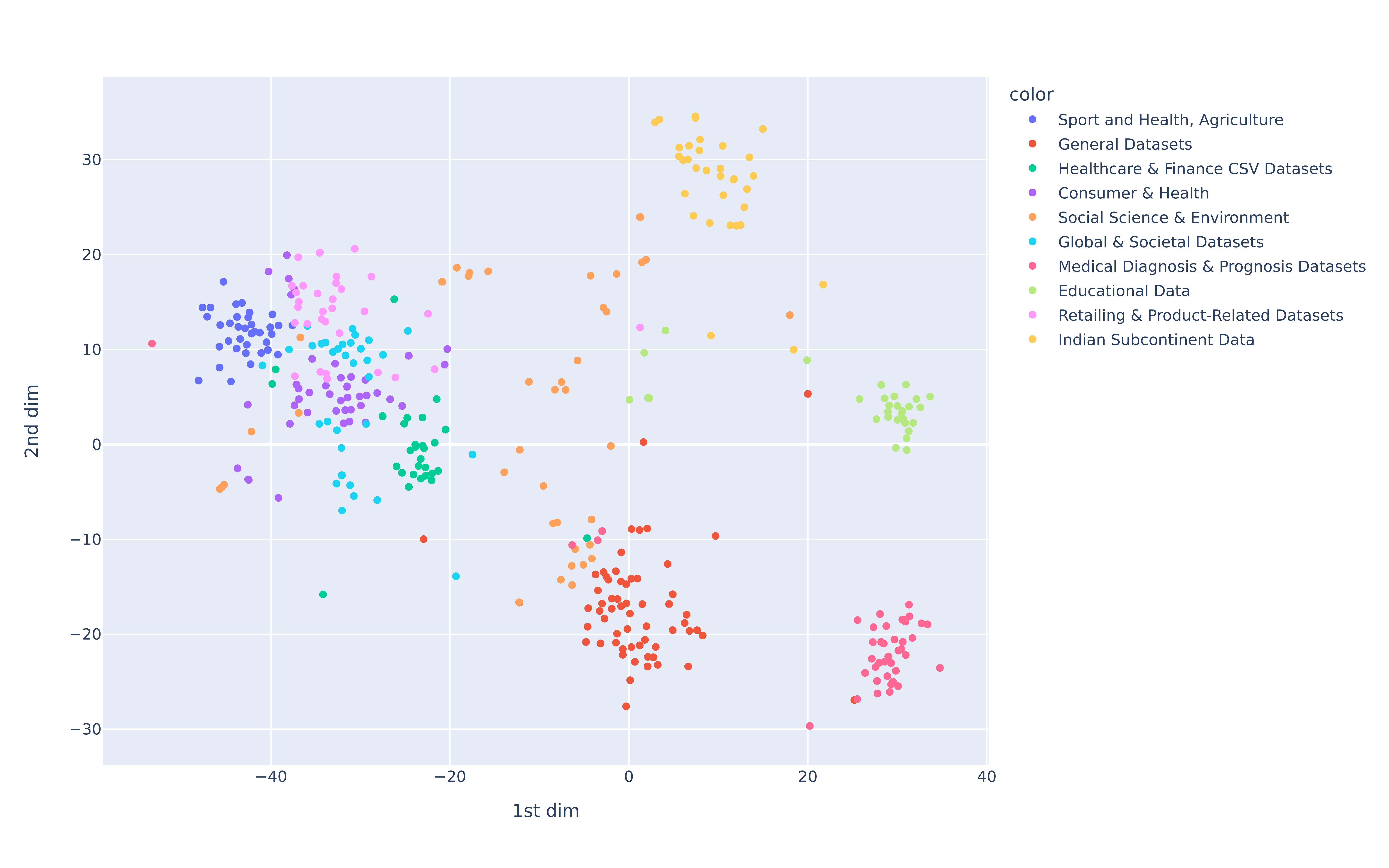}
    \caption{TSNE representation of top 10 domains clustered by $k$-means algorithm. Domain names are manually labeled by human by looking at the cluster keywords. \label{fig:domain kmean top 10}}
    
\end{figure}

 Kaggle is a leading platform for data science, offering diverse datasets across fields like healthcare and finance. A key challenge in using Kaggle data is its varied formats, duplicated information, and noisy data, including numerical, categorical, and unstructured data like text or time series. Our initial step filters Kaggle datasets by file type and quality, prioritizing those with a usability rating of 8/10 or higher. This process identified 43,514 potential datasets. Further cleaning based on data type, missing values, and unique categorical values narrowed it down to 1,435 high-quality tables containing only numerical and categorical data. We developed two variations of the Kaggle dataset. The first uses a random split to create training, validation, and test subsets. The second uses domain-based splits, clustering table names BERT embeddings into 100 groups with $k$-mean and then randomly assigning these clusters to the subsets. Figure \ref{fig:domain kmean top 10} shows a TSNE representation of the top 10 domains clustered by $k$-mean, illustrating clear separation in the projected space. This domain-based split aims to validate model transferability across domains. Detailed statistics are in Table \ref{tab:statistics main}, with further details on the cleaning and splitting process in the Appendix.

\begin{table}[]
\centering
\caption{\label{tab:statistics main}Statistics about the datasets}
\begin{tabular}{llllll}
\hline
\textbf{Dataset} & \textbf{Raw tables} & \textbf{\begin{tabular}[c]{@{}l@{}}Cleaned tables \\ train/val/test\end{tabular}} & \textbf{Avg. \# columns} & \textbf{Avg. \# rows}  & \textbf{License}\\ \hline
Kaggle Random &    43514   & 1148/143/144      & 8.37      & 224.8   &  Kaggle         \\ \hline
Kaggle Domains &    43514   & 969/218/248      & 8.37      & 224.8   & Kaggle           \\ \hline
GitTables &   1M  & 1006/126/126      & 9.51      & 1112.68   & CC BY 4.0          \\ \hline
\end{tabular}
\end{table}

\subsection{GitTables Dataset}

 The GitTables corpus \cite{hulsebos2023gittables}, a large-scale collection of relational tables from CSV files on GitHub, supports the development of table representation models and applications in data management and analysis. It includes one million raw tables, many containing unstructured data like text (e.g., product reviews) or time series (e.g., server bandwidth). This requires significant data cleaning for foundational model building. After filtering for tables with only structured data, 1,258 tables remain, split into train/validation/test datasets as shown in Table \ref{tab:statistics main}.
\section{The TabularFM Framework}
TabularFM is an end-to-end framework for studying Tabular Foundational Models. In the preprocessing phase, it supports configurable data manipulation, including automated cleaning, metadata generation, and data splitting and shuffling. During training, it offers various tabular generative models with corresponding data transformation methods. For evaluation, it provides metrics to assess data synthesizer performance and transferability. Basic concepts, data transformation, experimental settings, and evaluation are briefly introduced here; detailed information is in the Appendices.

\subsection{Learning Methods}

\paragraph{Conditional Tabular GAN (CTGAN).} CTGAN, proposed by \cite{xu2019modeling}, is a conditional GAN-based model specialized for generating tabular data. It handles mixed numerical and categorical data through datatype-specific transformations. To address mode collapse and imbalance, CTGAN employs a PacGAN-style approach \cite{NEURIPS2018_288cc0ff} and a training-by-sampling strategy, incorporating conditional vectors and adjusting the generator loss. It is trained using the WGAN method with gradient penalty \cite{NIPS2017_892c3b1c}.

\paragraph{Tabular Variational Autoencoder (TVAE).} While proposing CTGAN, \cite{xu2019modeling} also introduce a Variational Autoencoder designed for tabular data generation. The architecture mainly follow conventions and the model is optimized by using evidence lower-bound (ELBO) loss.

\paragraph{Shared Tabular Variational Autoencoder (STVAE).}
We notice that TVAE uses a trainable parameter for standard deviations of columns, limiting dataset transferability during pretraining. To solve this, we remove this parameter and directly optimize numerical values using the ELBO loss function, naming this model Shared TVAE (STVAE).

\paragraph{Shared Tabular Variational Autoencoder with Metadata (STVAEM).}
To enhance transferability, we add signature information to tables, which is the same for a dataset. We use embeddings of column names from a pretrained language model as this signature information and concatenate it with the input data, calling this model STVAEM.

\paragraph{Generation of Realistic Tabular Data (GReaT).}
Transformer-based models aim to maximize the probability of predicting the next token based on previous tokens, like auto-regressive language models \cite{Jelinek1980InterpolatedEO, NIPS2000_728f206c}. Thus, any pretrained generative language model can be used. GReaT \cite{borisov2023language} initializes training from large pretrained models to enhance tabular data generation through contextual representation. In this work, we use generative transformer-decoder LLM architectures \cite{radford2018improving, radford2019language, brown2020language}, specifically a distilled version of GPT, as the baseline model.

\subsection{Data Transformation}
Following \cite{xu2019modeling}, we utilized one-hot encoding for the categorical columns. For each numerical column, we fitted a mixture of Gaussians with a user-defined $K$ modes. Each numerical value was normalized by subtracting the mean value and dividing by four times the standard deviation of the corresponding predicted mode. This normalization ensures that the scales of the data across tables are ignored.  For transformer-based models \cite{borisov2023language}, no specific data normalization is needed except converting each data value $t_{i,j}$ into a sentence of the form: column name \text{ "is" } $t_{i,j}$. An example for a table with columns Age and Gender is ``\texttt{Age is 26 and Gender is M}". For a formal definition of the data transformation process, please refer to the Appendix.

% We deliver off-the-shelf transformation for supported learning methods. For methods such as CTGAN and TVAE, we follow their proposed transformation. Gaussian Mixture Model \cite{} is leveraged to calculate distributions of the numerical data, then, a mode is randomly sampled. As a result, each numerical column is represented by two kinds of information, a normalized value from the chosen mode, and a one-hot vector representing mode information. Categorical columns are represented by one-hot vectors. Afterwards, transformed data is concatenated in order.\\

\subsection{Experiment Settings}

Let $\pmb{\mathcal{D}} = \{\mathcal{D}_i\}$ represent all datasets across domains. We divide them into three types: $\pmb{\mathcal{D}}^{\text{pretrain}}$ (pretraining), $\pmb{\mathcal{D}}^\text{val}$ (validation), and $\pmb{\mathcal{D}}^\text{test}$ (test) datasets. Our goal is to study the transferability of tabular datasets. We train a model on $\pmb{\mathcal{D}}^{\text{pretrain}}$, then fine-tune it on each dataset in $\pmb{\mathcal{D}}^\text{val}$ and $\pmb{\mathcal{D}}^\text{test}$. Additionally, we train a separate model from scratch for each dataset. We compare the performance of fine-tuning against training from scratch.

To evaluate the performance of tabular generation, we compare the quality of synthetic data versus real data by two properties. We measure the distribution of columns data between synthetic and real data, called \textit{Column Shapes}. Additionally, correlation among columns is also computed, called \textit{Column Trends}. We then average the two metrics as \textit{Overall Score}. As a result, we compare the performance of synthetic data to that of real data based on those metrics.
% As a result, we compare the performance between ones of synthetic and real data based on those metrics. 

In Table \ref{tab:models main}, we summarize the statistics of the pretrained models. Due to space constraints, comprehensive information about data cleaning and preparation, hyperparameter tuning, additional details about model training and evaluation, and computing resources are thoroughly described in the Appendix.
\begin{table}[]
\centering
\caption{\label{tab:models main}Statistics about the pretrained models in our work}
\begin{tabular}{lrllll}
\hline
\textbf{Model} & \textbf{\# Parameters} & \textbf{Dataset} & \textbf{Method} & \textbf{Split} & \textbf{License}\\ \hline
ctgan\_gittables &    97,243,685   & GitTables      & CTGAN      & Random & MIT             \\ \hline
stvae\_gittables &    9,315,214   & GitTables      & STVAE      & Random              & MIT\\ \hline
great\_gittables &    81,912,576   & GitTables      & GREAT     & Random              & MIT\\ \hline
ctgan\_kg\_random &    97,243,685   & Kaggle      & CTGAN      & Random              & Kaggle\\ \hline
ctgan\_kg\_domain  &   97,243,685  & Kaggle      & CTGAN      & By domain             & Kaggle\\ \hline
stvae\_kg\_random &    9,315,214   & Kaggle      & STVAE      & Random              & Kaggle\\ \hline
stvae\_kg\_domain  &   9,315,214  & Kaggle      & STVAE      & By domain             & Kaggle\\ \hline
great\_kg\_random &    81,912,576   & Kaggle      & GREAT      & Random              & Kaggle\\ \hline
great\_kg\_domain  &   81,912,576  & Kaggle      & GREAT      & By domain             &Kaggle\\ \hline
\end{tabular}
\end{table}

\section{Benchmark Results and Discussion}
\label{sec:benchmark results}
In this section, we discuss experimental results using various data splits from both the Kaggle and GitTables datasets. These results are presented as standard benchmark outcomes for evaluating foundation models for tabular data. We have created corresponding leaderboards, which will be hosted on the Papers with Code platform\footnote{\url{https://tabularfm.github.io}}.
\subsection{Model Transferability Comparison}
\label{sec:Model Transferability Comparison}
The results of the models on three datasets—Kaggle with random split, GitTables, and Kaggle split by domains—are presented in Table \ref{tab:transferability}. All models were trained on the training subsets and validated on the test subsets, comprising 144, 126, and 248 test tables respectively. Evaluation metrics, including column shape, column pair trends, and the overall average, were averaged across all test tables. A \textit{Mann-Whitney U} test was conducted to compare the distributions of results from training from scratch and pretrained models, with the null hypothesis that they are equal and the alternative hypothesis that they are not.

We present results for each learning method with and without pretraining, denoted as TVAE and TVAE Pretrained, respectively. Pretrained models of TVAE and CTGAN consistently outperform those trained from scratch by about 10 points on average. Adding meta-information did not enhance transferability, posing a future challenge of identifying optimal meta-information. Results are consistent across domain-based and random splits, with slightly worse performance in domain-based splits. These findings highlight the need for better data splitting methods to ensure maximum independence, which we plan to explore in future work.

%In Table \ref{tab:transferability random gittables}, we present the results from the GitTables dataset. Overall, the pretrained models outperform models trained from scratch by 10 points. Unlike the results with the Kaggle dataset, the STVAE pretrained model performed best, followed by the CTGAN pretrained model. The results from both benchmark datasets confirm that state-of-the-art representation learning methods for tabular data can transfer patterns between tables effectively, and the benefit of using pretrained models is validated by these experiments.

\begin{table}[]
\centering
\caption{Transferability of pretrained models on Kaggle and GitTables dataset. The reported metrics (larger is better) and the standard deviation are shown in brackets.\label{tab:transferability}}
\begin{tabular}{llllll}
\toprule
\textbf{Data/Split}        & \textbf{Method} & \textbf{Shape} & \textbf{Trends} & \textbf{Overall} & \textbf{p-value}        \\ 

\midrule

Kaggle Random
& GREAT Pretrained         & \textbf{0.73} (0.16)           & 0.50 (0.25)                 & \textbf{0.61} (0.18)      & 0.9 \\ 
                            & GREAT      & 0.73 (0.16)           & 0.50 (0.25)                 & \textbf{0.61} (0.18)      &                         \\
% \cline{2-6}
\cmidrule(r){2-6}
& CTGAN Pretrained         & 0.70 (0.16)           & 0.49 (0.26)                 & 0.59 (0.19)      & 3.3e-8 \\ 
                            & CTGAN      & 0.48 (0.17)           & 0.47 (0.25)                 & 0.48 (0.17)      &                         \\
% \cline{2-6}
\cmidrule(r){2-6}
                            & TVAE Pretrained           & 0.47 (0.13)           & 0.49 (0.27)                 & 0.48 (0.17)      & 0.0039 \\

                            & TVAE         & 0.55 (0.13)           & \textbf{0.52} (0.26)                 & 0.54 (0.18)      &                          \\
% \cline{2-6}
\cmidrule(r){2-6}
                            & STVAE Pretrained  & 0.54 (0.16)           & 0.45 (0.28)                 & 0.50 (0.19)      & 8.5e-4 \\

                            & STVAE & 0.44 (0.14)           & 0.41 (0.28)                 & 0.43 (0.17)      &                          \\
% \cline{2-6}
\cmidrule(r){2-6}
                            & STVAEM Pretrained& 0.43 (0.17)           & 0.40 (0.24)                 & 0.42 (0.16)      &     0.02                      \\
                             & STVAEM& 0.35 (0.15)           & 0.40 (0.27)                 & 0.37 (0.16)      &                          \\
                           
\midrule
GitTables
& GREAT Pretrained         & 0.68 (0.22)           & 0.56 (0.27)                 & 0.61 (0.19)      & 0.46 \\ 
                            & GREAT      & \textbf{0.71} (0.20)           & 0.58 (0.25)                 & \textbf{0.63} (0.17)      &                         \\
% \cline{2-6}
\cmidrule(r){2-6}

& CTGAN Pretrained         & 0.67 (0.12)           & 0.55 (0.24)                 & 0.61 (0.14)      & 3.4e-5 \\ 
                            & CTGAN      & 0.50 (0.18)           & 0.58 (0.24)                 & 0.53 (0.15)      &                         \\
% \cline{2-6}
\cmidrule(r){2-6}
                            & TVAE Pretrained           & 0.45 (0.13)           & 0.49 (0.27)                 & 0.47 (0.15)      & 0.0006 \\

                            & TVAE         & 0.54 (0.12)           & 0.53 (0.27)                 & 0.54 (0.13)      &                          \\
% \cline{2-6}
\cmidrule(r){2-6}
                            & STVAE Pretrained  & 0.55 (0.12)           & \textbf{0.59} (0.27)                 & 0.57 (0.13)      & 1.2e-7 \\

                            & STVAE & 0.45 (0.12)           & 0.50 (0.27)                 & 0.47 (0.13)      &                          \\
                           
\midrule
Kaggle Domain
                        & GREAT Pretrained         & 0.72 (0.14)           & \textbf{0.56} (0.23)                 & \textbf{0.64} (0.16)      & 0.7 \\ 
                            & GREAT      & \textbf{0.73} (0.15)           & \textbf{0.56} (0.23)                 & \textbf{0.64} (0.16)      &                         \\
% \cline{2-6}
\cmidrule(r){2-6}
                        & CTGAN Pretrained         & 0.69(0.12)           & 0.53 (0.24)                 & 0.62 (0.15)      & 1.4e-14 \\ 
                            & CTGAN      & 0.49 (0.16)           & 0.53 (0.22)                 & 0.51 (0.14)      &                         \\
% \cline{2-6}
\cmidrule(r){2-6}
                            & TVAE Pretrained           & 0.39 (0.13)           & 0.45 (0.27)                 & 0.42 (0.17)      & 9.7e-10 \\

                            & TVAE         & 0.53 (0.13)           & 0.39 (0.26)                 & 0.51 (0.18)      &                          \\
% \cline{2-6}
\cmidrule(r){2-6}
                            & STVAE Pretrained  & 0.48 (0.13)           & 0.43 (0.25)                 & 0.46 (0.15)      & 0.33 \\

                            & STVAE & 0.42 (0.12)           & 0.46 (0.27)                 & 0.44 (0.15)      &                          \\
% \cline{2-6}
\cmidrule(r){2-6}
                            & STVAEM Pretrained& 0.43 (0.11)           & 0.39 (0.22)                 & 0.41 (0.13)      &     0.02 \\
                             & STVAEM& 0.34 (0.13)           & 0.45 (0.26)                 & 0.40 (0.14)      &                          \\
                           
\bottomrule
\end{tabular}
\end{table}
\label{sec:random split}

The results for transformer models show that transformers pretrained on textual data (GREAT) achieve the best results on the benchmark, while further pretraining with train splits containing tabular data (GREAT Pretrained) slightly decreases performance. There are a few potential reasons behind this behavior. First, methods such as CTGAN and TVAE are not column permutation invariant, whereas the attention mechanism adopted in transformers is, allowing for more effective representation learning. Second, transformers are based on GPT-2, which was trained on a larger corpus of textual data. Even though GPT-2 was not trained on tabular data, it has acquired general knowledge that can be transferred to tabular data. Therefore, a future challenge is to create larger training tabular datasets to pretrain transformers in a way that outperforms current text-based models.

\begin{figure}[h]
    \centering
    \includegraphics[width=1.0\textwidth]{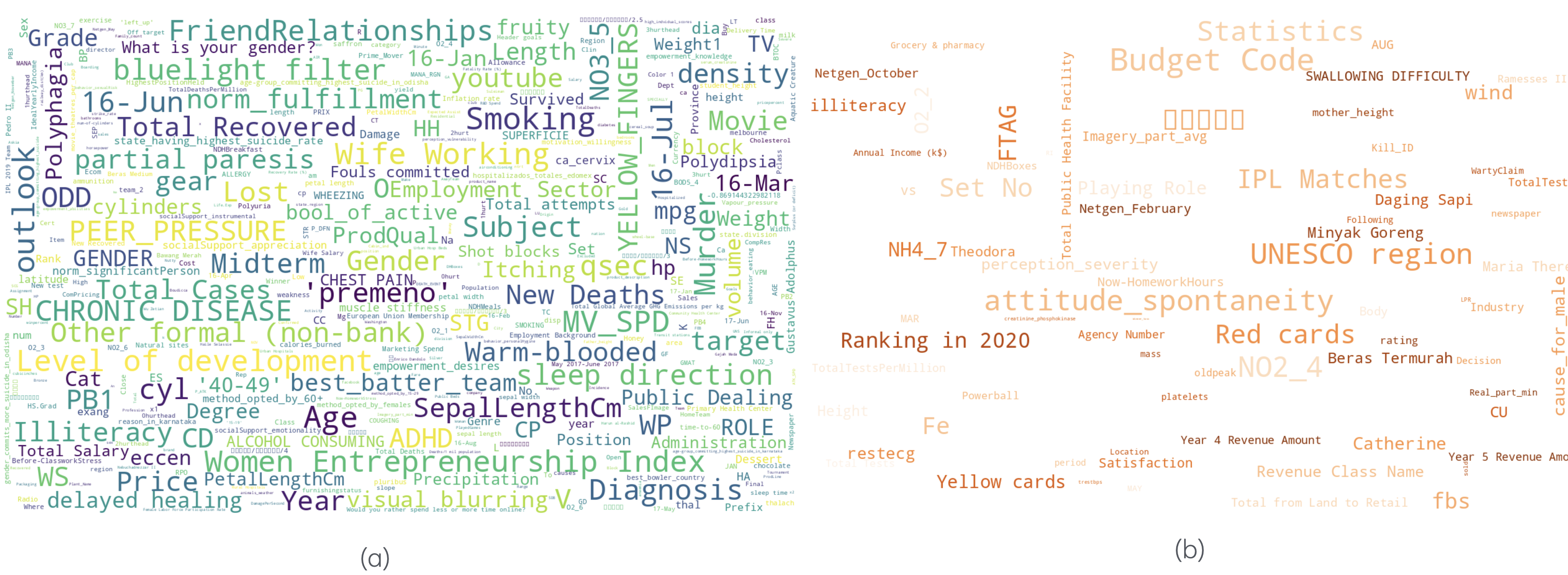}
    \caption{Wordclouds demonstrate columns with high transferability (a) and low transferability (b).}
    \label{fig:wordcloud}
\end{figure}
\begin{figure}[h]
    \centering
    \includegraphics[width=1.0\textwidth]{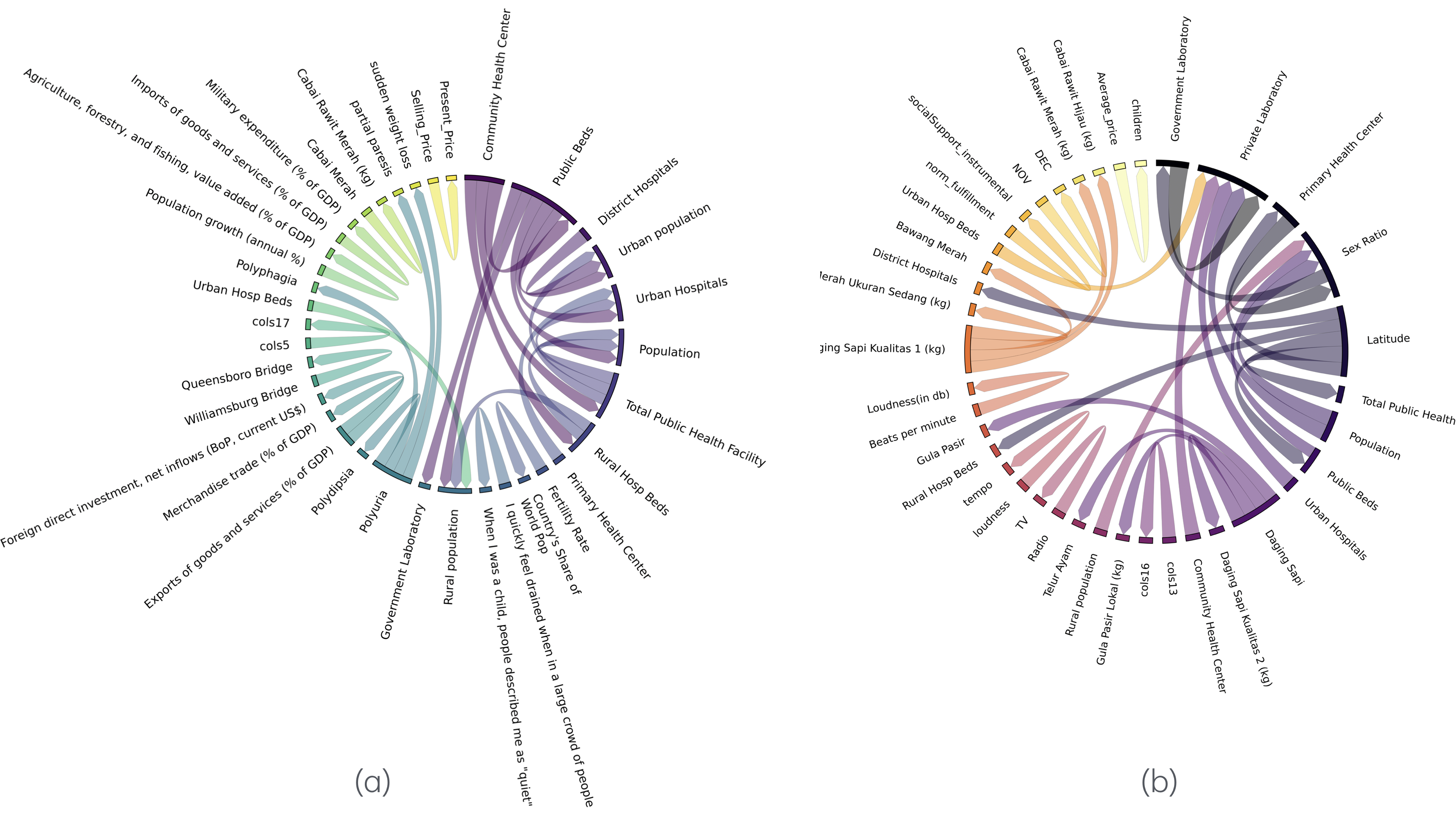}
    \caption{\label{fig:correlation} The plots illustrate the column pairs with the most significant differences between the pair trends of the pretrained models and the model trained from scratch in (a), and vice versa in (b).}
    
\end{figure}
\subsection{Transferability Analysis}
To understand the key factors driving the transferability results discussed in the previous section, we conducted further analysis. This analysis illuminates the types of knowledge that pretrained models can and cannot transfer. Figure \ref{fig:wordcloud} presents wordclouds of the column names, where the size of each word corresponds to the absolute value of the difference between the performance of the pretrained models and the model trained from scratch. Figure \ref{fig:wordcloud}.a displays columns where the pretrained models outperform the models trained from scratch, while Figure \ref{fig:wordcloud}.b shows the opposite. We observe that highly general knowledge such as \textit{Age}, \textit{Gender}, \textit{Chronic Disease} is transferable, whereas specific knowledge such as \textit{Unesco region}, \textit{Budget Code}, \textit{Red Cards} is not, which is logical.
\label{sec:domains}

To understand how pretrained models utilize general correlation patterns in data and transfer that knowledge across datasets, Figure \ref{fig:correlation} illustrates the links between columns where the difference between the pair trends of the pretrained models and the models trained from scratch is highest (Figure \ref{fig:correlation}.a) and vice versa (Figure \ref{fig:correlation}.b). We observe that general knowledge about the correlation between hospital beds and urban population or the link between weight loss and polydipsia via polyuria transfers well through pretraining. This finding is noteworthy because the models were pretrained solely by observing the data distribution, yet they could detect and transfer meaningful correlated patterns across datasets. Our analysis confirms that, despite tabular data being noisier than other types of data, specific types of knowledge can be transferred across data. Thus, developing foundational models for tabular data is a promising research direction.
\begin{figure}[h]
    \centering
    \includegraphics[width=1.0\textwidth]{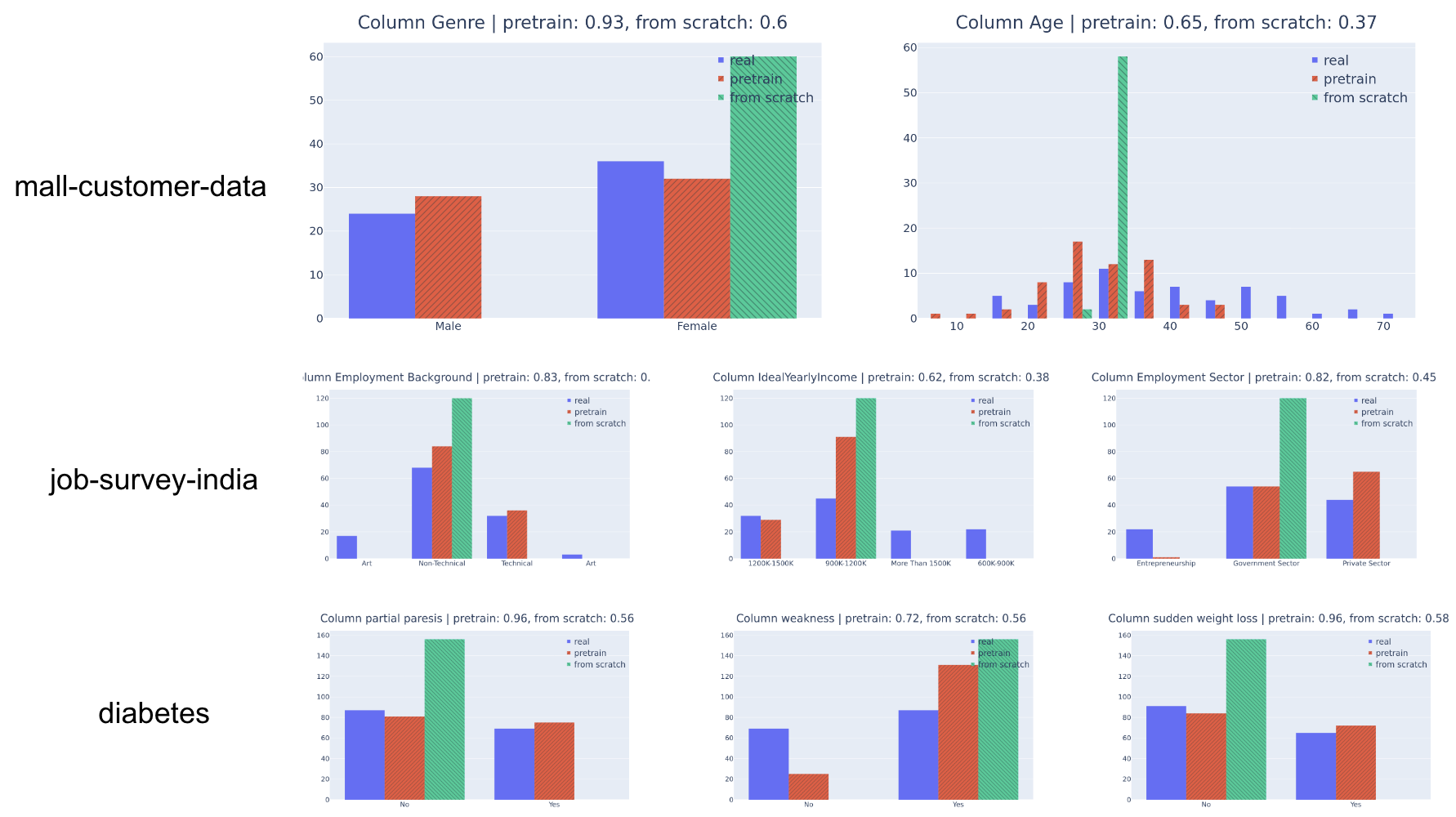}
    \caption{\label{fig:shape} Column shape comparison of STVAE Pretrained prediction  and STVAE from scratch.}
\end{figure}

\begin{figure}[h]
    \centering
    \includegraphics[width=1.0\textwidth]{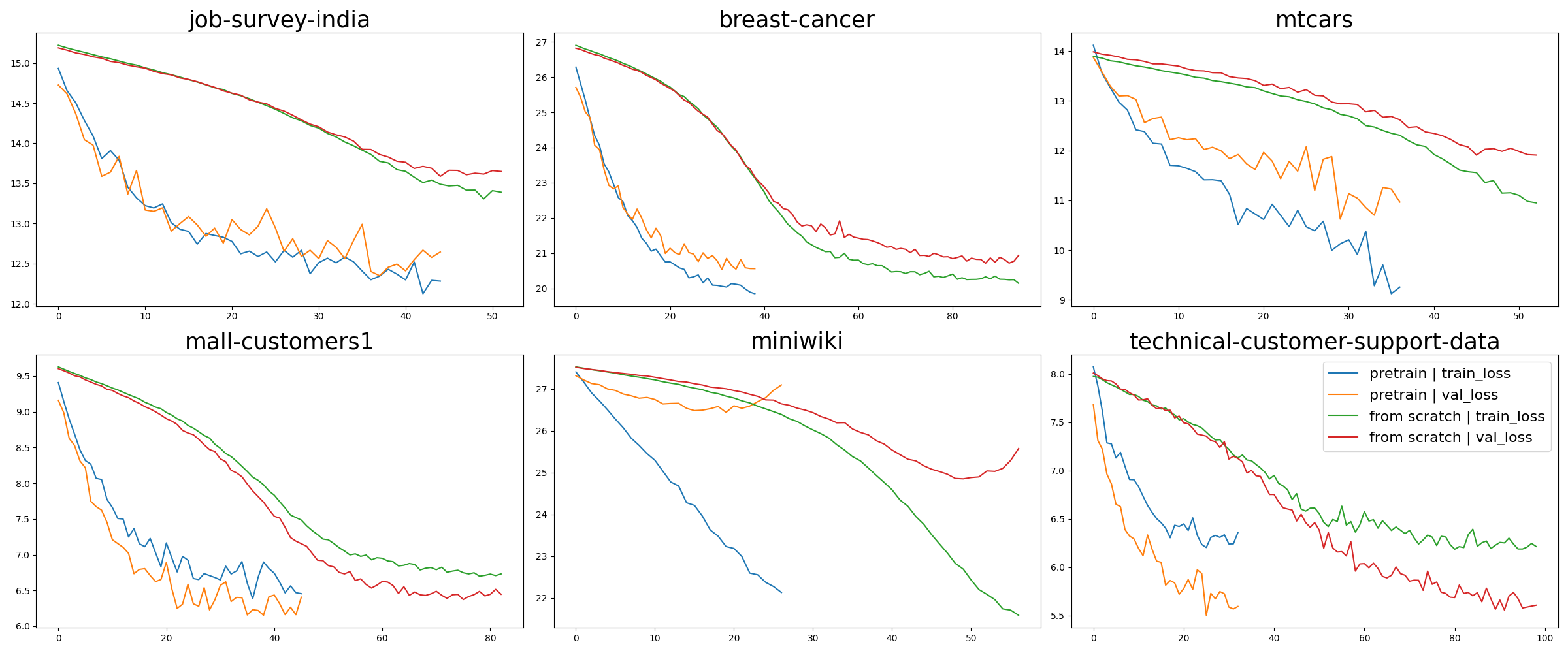}
    \caption{Training and validation loss of STVAE models when they are pretrained versus trained from scratch. Pretrained STVAE converge faster toward more optimal solutions.\label{fig:loss}}    
\end{figure}

In Figure \ref{fig:shape}, the histogram compares synthesized data with ground truth data. It reveals that the STVAE trained from scratch primarily predicts the mode of the column, neglecting the distribution tails. In contrast, the pretrained STVAE effectively captures the tails of the distribution. This difference is reflected in the faster learning curve shown in Figure \ref{fig:loss}, where the benefits of pretraining include quicker and better convergence. These findings are promising and confirm that transferability between tabular datasets is feasible.

\section{Related Work}

Learning representation for tabular data has seen significant advancements with various approaches aiming to enhance performance, interpretability, and efficiency.  

\textit{Generative Adversarial Network-based methods:}
Generative adversarial networks (GANs) have been explored for tabular data with models like TGAN \citep{xu2018synthesizing} and CTGAN \citep{xu2019modeling}. TGAN focuses on generating high-quality synthetic tables with mixed discrete and continuous variables, while CTGAN uses conditional GANs to address the challenges of modeling imbalanced and multi-modal tabular data. Both models demonstrate the potential of GANs to generate realistic synthetic tabular data, facilitating privacy-preserving data generation and enhancing data availability for training robust machine learning models.

% \textit{Customized deep-learning based approaches:}
\textit{Distribution-agnostic Deep Learning Approaches:}
\citep{Joffe_2021} proposes a CNN-based architecture for transferring patterns from tabular data by treating it as a 2D image, with a min-max scaling method for numerical columns to address scaling issues. The CNN learns spatial relations between columns, but since column order doesn't matter in tabular data, random shuffling of columns during training is required. \citep{iida-etal-2021-tabbie} introduces TABBIE, which uses two transformers to independently encode rows and columns, pooling their representations at each layer. Like training BERT by masking a random word, it corrupts a random cell in the input table and trains the model to predict the corrupted cells. This approach ignores the challenges associated with tabular data, such as numeric and categorical data normalization and column order invariance. TURL in \citep{Deng_2021} considers entity embedding using a transformer encoder for relational web tables. Each entity is represented with properties, and the data is kept in a relational web-table format. Similar to TABBIE, TURL first serializes the tables into sequences and then trains the transformer-based model with a masking objective similar to training BERT. The proposed method only works for clean data and cannot learn important patterns beyond the clean dataset used in the experiments. UNITABE \citep{yang2024unitabe} consists of a transformer encoder and a shallow LSTM decoder to leverage the relationship between column names and values. To handle varying data types, the authors propose a feature processor that considers each cell in a table as a key-value pair, representing the column name and the corresponding column value. The work \citep{arik1908tabnet} proposes TabNet -- a deep learning architecture for tabular data that uses sequential attention to select features at each decision step. This approach enables both local and global interpretability, as well as efficient learning by focusing on the most salient features. Its ability to handle raw tabular data without preprocessing and its application of unsupervised pretraining for improved performance show its strengths in handling diverse tabular datasets with high interpretability. Moreover, the paper \citep{nam2023stunt} proposes STUNT to improve performance in few-shot learning scenarios. STUNT applies meta-learning to generalize knowledge from these self-generated tasks by generating pseudo-labels from unlabeled data through $k$-means clustering of randomly chosen column subsets. TabPFN \citep{hollmann2022tabpfn} is a transformer-based model that solves supervised classification problems on small tabular datasets without hyperparameter tuning. This approach highlights the potential of pretrained models for efficient and accurate tabular data classification, setting a precedent for rapid and effective tabular data processing. TransTab \citep{wang2022transtab} is a method for learning transferable representations across tables with varying structures. TransTab allows pretraining on multiple distinct tables and fine-tuning on target datasets by converting table rows into generalizable embedding vectors and employing a gated transformer model that integrates column descriptions and cell values. This approach addresses the challenge of maintaining model performance across tables with different columns and structures, showing the versatility of transformer-based models for tabular data. Furthermore, the work \citep{padhi2021tabular} extends the application of transformers to tabular time series data. They demonstrated on both synthetic and real-world datasets by leveraging hierarchical structures of these models in tabular time series data to improve performance in downstream tasks like classification and regression. This work highlights the adaptability of transformer models to various tabular data formats, including time series.

\textit{Pretrained Large Language Model Adaptation:} 
Several authors propose methods to adapt pretrained large language models, such as LLaMA and GPT, to tabular data \citep{zhang2023foundation, hegselmann2023tabllm}. Using LLaMA, TabFM in \citep{zhang2023foundation} is trained with 115 public datasets by employing generative modeling of rows encoded as text along with task and column descriptions, incorporating additional loss for feature reconstruction. TabLLM in \citep{hegselmann2023tabllm} fine-tunes the T0 model by transforming a table into a natural text representation using serialization methods. LIFT \citep{dinh2022lift} is a language-interfaced fine-tuning method that fine-tunes GPT-based architectures by converting labeled samples into sentences with a fixed template then fine-tuning LLMs with sentence datasets.

All these approaches collectively highlight the advancements in leveraging transformer models and GANs for tabular data. Our framework, TabularFM, builds upon these foundations by introducing state-of-the-art methods tailored specifically for tabular data, including GANs, VAEs, and Transformers. By using an extensive collection of cleaned tabular datasets and offering pretrained models along with benchmarking tools, TabularFM aims to advance the field significantly, providing a comprehensive solution for developing and evaluating foundational models for tabular data. 

\section{Limitations, Conclusions  and Future Work}
\label{sec:conclusion, limitation and future work}
In this work, we have preprocessed and curated about a million of tables to build and benchmark tabular foundation models. Our comparative study and detailed analysis of the benchmark results demonstrate the advantages of pretraining, providing insights into how pretraining transfers general knowledge across tabular datasets. These findings confirm the potential applications of tabular FMs. However, the challenges in building effective tabular FMs remain an open question. Our work offers a framework for rapidly benchmarking new models in the future.

Our study is limited to three types of neural learning architectures and focuses on numerical and categorical data. For future work, we plan to extend our framework to encompass a more diverse range of data types stored in tabular format and to train larger models on more extensive datasets.    

\newpage

\bibliographystyle{unsrtnat}
\bibliography{references}  

% Appendix
\appendix
\newpage

  \vbox{%
    \hsize\textwidth
    \linewidth\hsize
    \vskip 0.1in
  \hrule height 4pt
  \vskip 0.25in
  \vskip -5.5pt%
  \centering
    {\LARGE\bf{TabularFM: an Open Framework for Tabular Foundational Models \\
    Supplementary Material} \par}
      \vskip 0.29in
  \vskip -5.5pt
  \hrule height 1pt
  \vskip 0.09in%
    
  \vskip 0.2in
  }

\section{Released Pretrained Models}
We have released all the pretrained models for the GitTables datasets under the MIT license on Huggingface\footnote{https://huggingface.co}. For the models trained on Kaggle datasets, we offer the datasets and detailed documentation on the training steps, as some Kaggle tables have restricted licenses. A summary of the models' statistics can be found in Table \ref{tab:models}. By offering these pretrained models, we enable researchers and practitioners to utilize state-of-the-art techniques without the need for extensive computational resources or time-consuming training processes. 
\begin{table}[h]
\centering
\caption{\label{tab:models}Statistics about the pretrained models in our work}
\begin{tabular}{lrllll}
\hline
\textbf{Model} & \textbf{\# Parameters} & \textbf{Dataset} & \textbf{Method} & \textbf{Split} & \textbf{License}\\ \hline
ctgan\_gittables &    97,243,685   & GitTables      & CTGAN      & Random & MIT             \\ \hline
stvae\_gittables &    9,315,214   & GitTables      & STVAE      & Random              & MIT\\ \hline
great\_gittables &    81,912,576   & GitTables      & GREAT     & Random              & MIT\\ \hline
ctgan\_kg\_random &    97,243,685   & Kaggle      & CTGAN      & Random              & Kaggle\\ \hline
ctgan\_kg\_domain  &   97,243,685  & Kaggle      & CTGAN      & By domain             & Kaggle\\ \hline
stvae\_kg\_random &    9,315,214   & Kaggle      & STVAE      & Random              & Kaggle\\ \hline
stvae\_kg\_domain  &   9,315,214  & Kaggle      & STVAE      & By domain             & Kaggle\\ \hline
great\_kg\_random &    81,912,576   & Kaggle      & GREAT      & Random              & Kaggle\\ \hline
great\_kg\_domain  &   81,912,576  & Kaggle      & GREAT      & By domain             &Kaggle\\ \hline
\end{tabular}
\end{table}

\section{Hyperparameter Sensitivity Analysis and Settings}
\begin{figure}[h]
     \centering
     \includegraphics[width=1.0\textwidth]{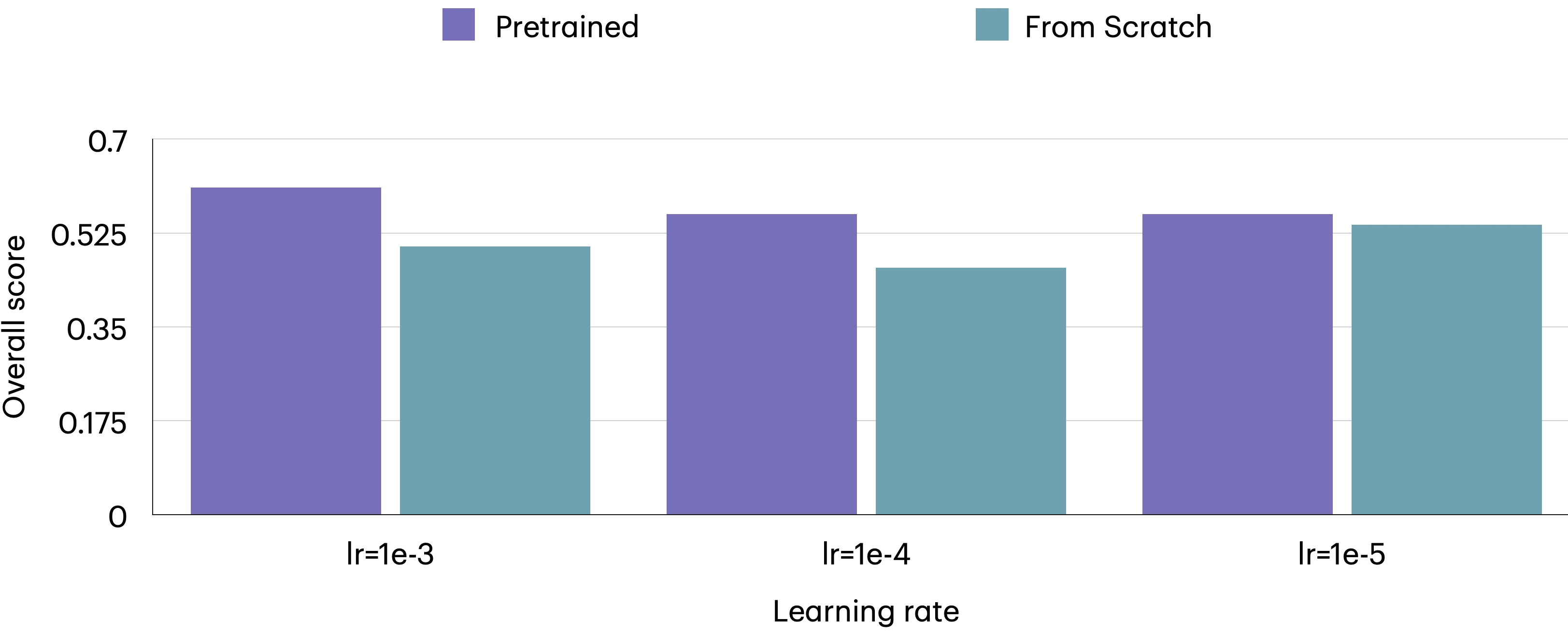}
     \caption{The impact of learning rate on the results of CTGAN with pretraining and training from scratch, the results were calculated on the validation tables of the random split of the Kaggle dataset.  \label{fig:learning_rate}}
 \end{figure}

\begin{figure}[h]
     \centering
     \includegraphics[width=1.0\textwidth]{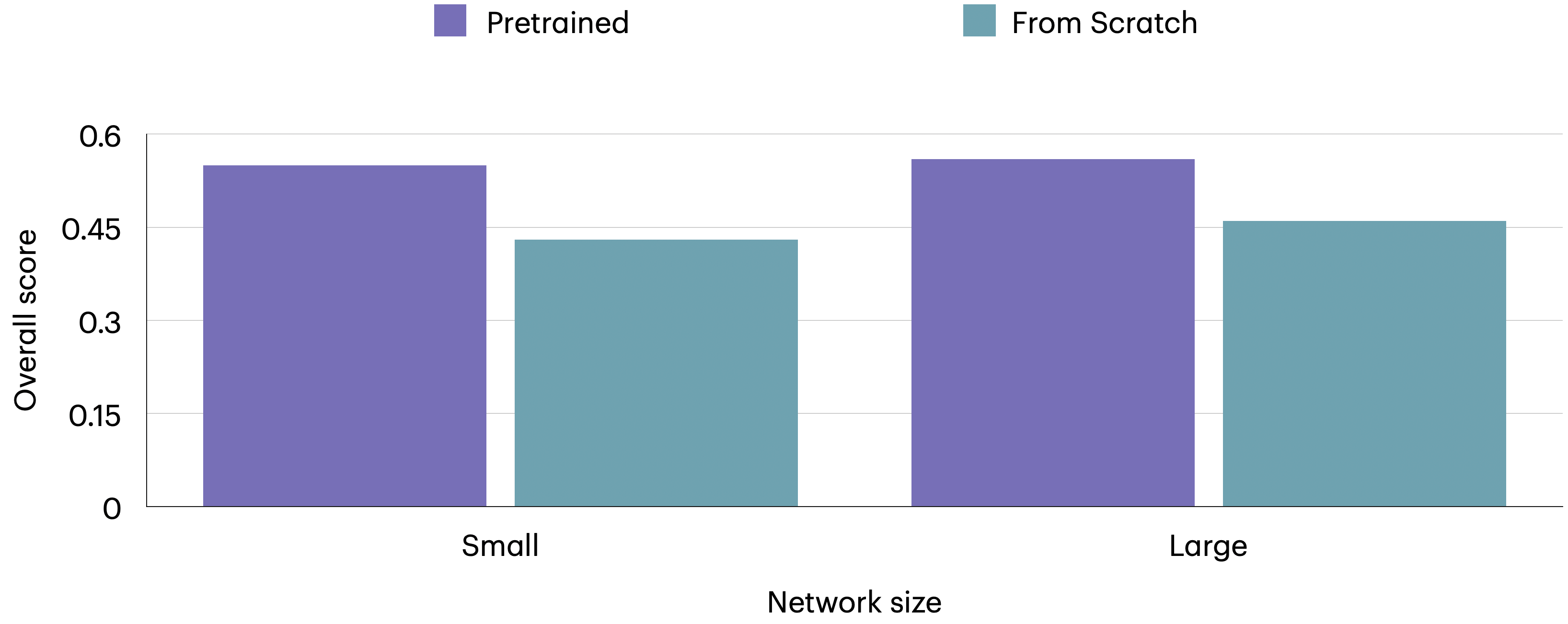}
     \caption{The impact of network size on the results of CTGAN with pretraining and training from scratch, the results were calculated on the validation tables of the random split of the Kaggle dataset.  \label{fig:network_size}}
 \end{figure}

The hyperparameters used in our experiments are summarized in Table \ref{tab:hpo}. To determine the optimal network size and learning rate for CTGAN and TVAE, we conducted pretraining with learning rates of 1e-3, 1e-4, 1e-5 and network sizes set to either small or normal ones. The performance results of the CTGAN model on the validation random split of the Kaggle dataset, showing the effects of varying network size and learning rate, are presented in Figure \ref{fig:learning_rate} and Figure \ref{fig:network_size} respectively. With different network sizes and learning rates, the performance of the pretrained model consistently surpassed that of models trained from scratch. Besides, when the model size is reduced, there is a slight decline in performance compared to larger models. Thus, increasing the model size is a potential strategy to further enhance results. For GREAT, we used the recommended default network and learning rate. 

\begin{figure}[h]
    \centering
    \includegraphics[width=1.0\textwidth]{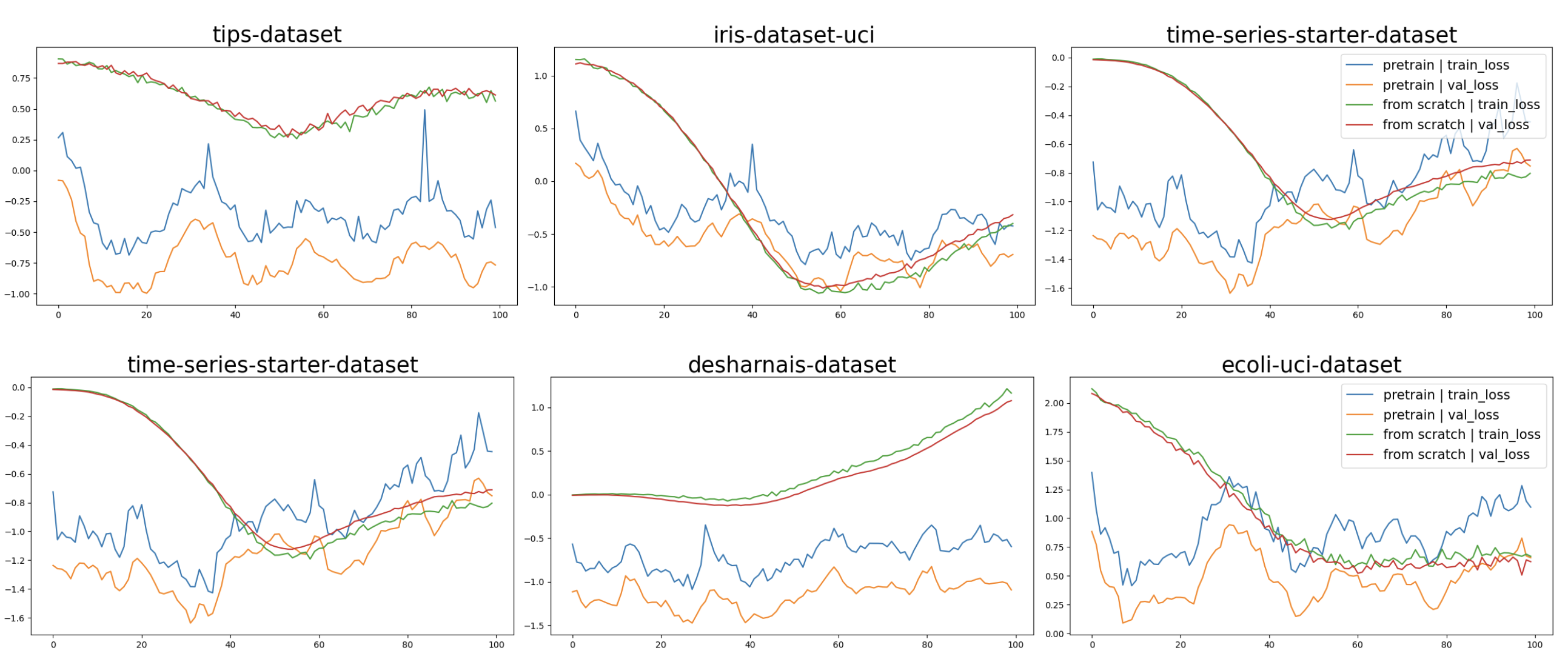}
    \caption{Training and validation loss of the generator of the CTGAN models when they are pretrained and finetuned on the data versus trained from scratch.}
    \label{fig:gloss}
\end{figure}

\section{Computing Resource}
We conducted all the experiments on a machine equipped with 4 CPUs, 50GB of memory, and one A100 GPU. Each pretraining task was configured to run for a maximum of 2 days or 500 epochs, whichever was reached first. Fine-tuning and training from scratch require less computing power and typically complete within 24 hours.

\section{Transferability Results Analysis}
In this section, we discuss additional analysis on transferability. Figure \ref{fig:gloss} illustrates the training and validation loss of the generator of the CTGAN models when they are pretrained and finetuned on the data versus trained from scratch. We can see the pretrained generators provide much better initial prediction and tends to converge faster to better solutions.

\begin{figure}[h]
    \centering
    \includegraphics[width=1.0\textwidth]{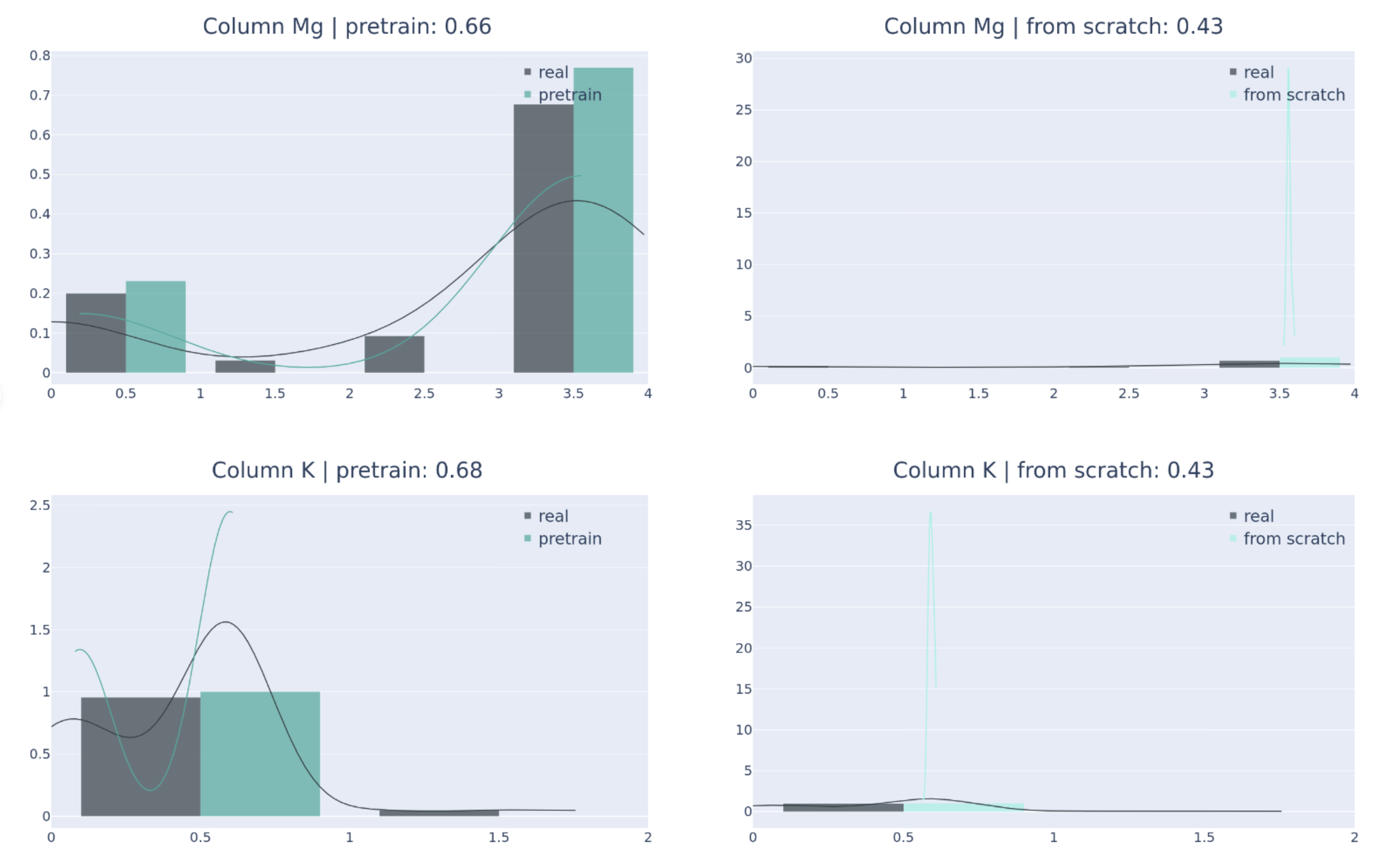}
    \caption{Column shape of pre-trained STVAE models and STAVE models trained from scratch for the table composition of glass.}
    \label{fig:column shape compition of glass}
\end{figure}

Figure \ref{fig:column shape compition of glass} illustrates the performance of pre-trained STVAE models compared to STVAE models trained from scratch for the table composition of glass. The distribution predicted by the pre-trained models closely aligns with the ground truth data for each column. In contrast, the STVAE models trained from scratch tend to overfit to the mode of the distribution, likely due to the limited data available.

\section{Supported Learning Methods}

We briefly describe the technical details of the learning methods used in this paper.

\subsection{Conditional Tabular GAN (CTGAN).}

CTGAN is a conditional GAN-based model for tabular data proposed by \cite{xu2019modeling}. Particularly, a table row $\mathbf{r}$ in the dataset $\mathcal{D}$, i.e. $\mathbf{r} \in \mathcal{D} \sim \mathbb{P}$ where $\mathbb{P}$ is the distribution of real dataset $\mathcal{D}$. To generate a synthetic data $\hat{\mathbf{r}}$, a generator $\mathcal{G}(\mathbf{z})$, where $\mathbf{z} \sim \mathcal{N}(\mathbf{0}, \mathbf{1})$ is learned, formally. 

\begin{equation}
    \hat{\mathbf{r}} \sim \mathbb{P}_{\mathcal{G}}(\mathbf{z})
\end{equation}

In the setting of conventional GAN, to valdiate the performance of the generator, a discriminator, or a critic score $\mathcal{C}$ is leveraged.
\begin{equation}
    \mathcal{C}(\mathbb{P}, \mathbb{P}_{\mathcal{G}})
\end{equation}

However, GAN may encounter mode collapse in case of imbalance datasets. To address this problem, CTGAN add a condition to make the synthetic data attached to a pre-defined category. Particularly, a specific category $k^* \in D_i$ is appointed during the generation, thus

\begin{equation}
\label{eq:sampling_rhat_from_conditionG}
    \hat{\mathbf{r}} \sim \mathbb{P}_{\mathcal{G}}(\mathbf{z} \vert D_i = k^*)    
\end{equation}

To accomplish this, CTGAN modelled the condition in \eqref{eq:sampling_rhat_from_conditionG} as conditional vector, i.e. $\textit{cond}$ and modify the generator loss to optimize it. Furthermore, training-by-sampling strategy is proposed to optimize CTGAN.

\paragraph{Conditional vector.} All categorical columns $D_1, D_2, \dots, D_{N_d}$ are leveraged to initiate conditional vectors. Suppose that a specific categorical value $k^*$ is chosen from a categorical column $D_{i^*}$, a one-hot vector $\mathbf{m}_i = [m_i^{(k)}], k=1, \dots, |D_i|$ is obtained as the mask, which is described as

\begin{equation}
    m_i^{(k)} = 
    \begin{cases}
            1, & \text{if}\ i=i^* \text{and} k = k^* \\
            0, & \text{otherwise}
    \end{cases}
\end{equation}

Hence, a conditional vector $cond$ is represented as 

\begin{equation}
    cond = \mathbf{m}_1 \oplus \mathbf{m}_2 \oplus \dots \oplus \mathbf{m}_{N_d}
\end{equation}

\paragraph{Generator loss.} It is obvious that conditional vector should be leveraged in the optimization. In fact, each mask one-hot vector $\mathbf{m}_i$ is constrained the generated data $\hat{\mathbf{d}}_i \in D_i$ to enforce it following the given condition. Hence, a cross-entropy loss between $\mathbf{m}_i$ and $\hat{\mathbf{d}}_i$ is supplemented into the existing generator loss of CTGAN to optimize the conditional vectors.

\paragraph{Training-by-sampling strategy.} We further clarify the sampling strategy with the involvement of conditional vectors to help optimizing the discriminator as follows
\begin{enumerate}
    \item Initiate original masked conditional vectors: setup all $\mathbf{m}_i = [m_i^{(k}]_{k=1,2,\dots, |D_i|}, i=1,2,\dots,N_d$.
    
    \item Randomly select a target categorical column: let $i^*$ be the index of chosen categorical column, $D_{i^*}$ is sampled with equal probability.
    
    \item Sample a target category of the selected categorical column: construct PMF over all categories of $D_{i^*}$ and randomly select a category $k^* \in D_{i^*}$.
    
    \item Adjust conditional vectors: modify the value $\mathbf{m}_{i^*}^{(k^*)} = 1$
    
    \item Finalize the conditional vectors: concatenate all masked vectors, $cond = \mathbf{m}_1 \oplus \mathbf{m}_2 \oplus \dots \oplus \mathbf{m}_{N_d}$
\end{enumerate}

Algorithm \ref{alg:batch_training_ctgan} describes training by batch procedure of CTGAN. The model are trained using the gradient penalty manner of WGAN \cite{NIPS2017_892c3b1c} and is further leveraged the PacGAN style to get rid of mode collapse \cite{NEURIPS2018_288cc0ff}.

\begin{algorithm}
    \caption{Training CTGAN by batch}

    \textbf{Input: } Dataset $\mathcal{D}$ with $\mathbf{T}_{train}$ transformed data, Generator $\mathcal{G}$ with paramters $\Phi_\mathcal{G}$, Critic $\mathcal{C}$ with parameters $\Phi_\mathcal{C}$, Optimizer $\mathcal{O}$ with learning rate $\eta$\\
    \textbf{Output: } \\
    
    \begin{algorithmic}[1]
        \State Initiate masks $\{[\mathbf{m}_i]_{i=1, \dots, N_d}\}_j$, \qquad for $1 \leq j \leq batch$ \label{state:start}
        
        \State Create conditional vectors $\{cond_j\}$, \qquad for $1 \leq j \leq batch$
    
        \State Sample $\{z_j\} \sim \mathcal{N}(0, \mathbf{I})$, \qquad for $1 \leq j \leq batch$
        
        \item[]

        % Generate synthetic data
        \State Generate $\hat{\mathbf{r}}_j \gets \mathcal{G}(z_j, cond_j)$, \qquad for $1 \leq j \leq batch$ \Comment{Generate synthetic data}

        % Sample real data
        \State Sample $\mathbf{r}_j \sim \text{Uniform}(\mathbf{T}_{train} | cond_j)$, \qquad for $1 \leq j \leq batch$ \Comment{Sample real data}

        \item[]
        
        % Apply Pac from PacGAN
        \State $cond_k^{(pac)} \gets cond_{k \times pac + 1} \oplus \dots \oplus cond_{k \times pac + pac}$, for $1 \leq k \leq batch / pac$ \Comment{Pac conditional vector} \label{state:generate_hat_r}

        \State $\hat{\mathbf{r}}_k^{(pac)} \gets \hat{\mathbf{r}}_{k \times pac + 1} \oplus \dots \oplus \hat{\mathbf{r}}_{k \times pac + pac}$, for $1 \leq k \leq batch / pac$ \Comment{Pac synthetic data}

        \State $\mathbf{r}_k^{(pac)} \gets \mathbf{r}_{k \times pac + 1} \oplus \dots \oplus \mathbf{r}_{k \times pac + pac}$, for $1 \leq k \leq batch / pac$ \Comment{Pac real data}

        \item[]

        % Caluclate Critic loss
        \State $\mathcal{L}_\mathcal{C} \gets \frac{1}{batch / pac} \sum_{k \in batch / pac}  \left( \mathcal{C}(\hat{\mathbf{r}}_k^{(pac)}, cond_k^{(pac)}) - \mathcal{C}(\mathbf{r}_k^{(pac)}, cond_k^{(pac)}) \right)$ \Comment{Calculate critic loss}

        \item[]

        % Gradient penalty
        \State Sample $\rho_1 \dots \rho_{batch/pac} \sim \text{Uniform}(0,1)$
        \State $\title{\mathbf{r}}_k^{(pac)} \gets \rho_k \hat{\mathbf{r}}_k^{(pac) + (1 - \rho_k) \mathbf{r}_k^{(pac)}}$, \qquad for $1 \leq k \leq batch / pac$
        \State $\mathcal{L}_{GP} \gets \frac{1}{batch / pac} \sum_{k \in batch / pac} \left( || \Delta_{\tilde{\mathbf{r}}_k^{(pac)}} \mathcal{C}(\tilde{\mathbf{r}}_k^{(pac)}, cond_k^{(pac))} ||_2 - 1\right)^2$ \Comment{Gradient penalty}

        \State Update gradient $\Phi_{\mathcal{C}}$ with $\mathcal{O}$ and $\eta$

        \item[]
        
        % Optimizer generator
        \State Generate $\hat{\mathbf{r}}_j$ following lines \ref{state:start} to \ref{state:generate_hat_r}.

        \State $\mathcal{L}_\mathcal{G} \gets -\frac{1}{batch / pac} \sum_{k \in batch / pac} \mathcal{C}(\hat{\mathbf{r}}_k^{(pac)}, cond_k^{(pac)}) + \frac{1}{batch} \sum_{j \in batch} \mathbb{CE}(\hat{\mathbf{d}}_{i^*, j}, \mathbf{m}_{i^*})$
        \State Update gradient $\Phi_{\mathcal{G}}$ with $\mathcal{O}$ and $\eta$
        
    \end{algorithmic}

\label{alg:batch_training_ctgan}
\end{algorithm}

\subsubsection{Architecture}

We denote $L$ as hidden dimension of fully connected layers in the architecture, $L_l$ represents the output dimension of that layer.

The generator architecture $\mathcal{G}(\mathbf{z}, \textit{cond})$ is formally described as

 \[~\left\{\begin{array}{@{}l@{}l@{}}
 \textbf{h}_0 = \mathbf{z} \oplus \textit{cond}\\[1em]
 
 \textbf{h}_{1} = \textbf{h}_{0} \oplus \text{ReLU}(\text{BN}(\text{FC}_{|\mathbf{h}_0|\rightarrow  L_0}(\textbf{h}_{0})))\\[1em]
 
\textbf{h}_{l+1} = \textbf{h}_{l} \oplus \text{ReLU}(\text{BN}(\text{FC}_{|\mathbf{h}_l| \rightarrow  L_{l}}(\textbf{h}_{l}))) & \qquad 1 \leq l \leq L \\[1em]
 
 \hat\alpha_i = \tanh(\text{FC}_{|\mathbf{h}_L| \rightarrow 1}(\textbf{h}_L)) & \qquad 1 \leq i \leq N_c \\[1em]
 
 \hat{\pmb\beta}_i = \text{gumbel}_{0.2}(\text{FC}_{|\mathbf{h}_L|  \rightarrow m_i}(\textbf{h}_L)) & \qquad 1 \leq i \leq N_c \\[1em]
 
 \hat{\mathbf{d}}_i = \text{gumbel}_{0.2}(\text{FC}_{|\mathbf{h}_L| \rightarrow |D_i|}(\textbf{h}_L)) & \qquad 1 \leq i \leq N_d
\end{array}\right.\]

and the architecture of critic $\mathcal{C}$ is defined as

\[~\left\{\begin{array}{@{}l@{}l@{}}
 \textbf{h}_0 = N_\text{pac} \times (\mathbf{r}_k \oplus \textit{cond}_k) \\[1em]
 
 \textbf{h}_1 = \text{drop}(\text{leaky}_{0.2}(\text{FC}_{|\mathbf{h}_0| \rightarrow L_0}(\mathbf{h}_0))\\[1em]
 \textbf{h}_{l+1} = \text{drop}(\text{leaky}(\text{FC}_{|\mathbf{h}_l| \rightarrow L_{l}}(\mathbf{h}_{l})) & \qquad 1 \leq l \leq L \\[1em]
 \mathcal{C} = \text{FC}_{|\mathbf{h}_L| \rightarrow 1}(\mathbf{h}_L)
\end{array}\right.\]

% \[~\left\{\begin{array}{@{}l@{}l@{}}
%  \textbf{h}_0 = \mathbf{r} \oplus \textit{cond}\\[1em]
%  \textbf{h}_1 = \text{drop}(\text{leaky}_{0.2}(\text{FC}_{|\mathbf{r}| + |\text{cond}| \rightarrow 128}(\mathbf{h}_0))\\[1em]
%  \textbf{h}_2 = \text{drop}(\text{leaky}_{0.2}(\text{FC}_{128 \rightarrow 256}(\mathbf{h}_1))\\[1em]
%  \textbf{h}_3 = \text{drop}(\text{leaky}_{0.2}(\text{FC}_{256 \rightarrow 256}(\mathbf{h}_2))\\[1em]
%  \textbf{h}_4 = \text{drop}(\text{leaky}_{0.2}(\text{FC}_{256 \rightarrow 512}(\mathbf{h}_3))\\[1em]
%  \mathcal{C} = \text{FC}_{256 \rightarrow 1}(\mathbf{h}_4)
% \end{array}\right.\]

% In particular, it is a conditional GAN-based method specializing for tabular data generation. CTGAN addresses the problem of mixing numerical and categorical data types by introducing a datatype-specific transformation. To deal with mode collapse and imbalance, the method also introduces training-by-sampling strategy.

\subsection{Tabular Variational Autoencoder (TVAE).} While proposing CTGAN, the authors also introduce a Variational Autoencoder designed for tabular data generation. Let $q_\phi(\mathbf{z}_j | \mathbf{r}_j)$ and $p_\theta(\mathbf{r}_j | \mathbf{z}_j)$ are the encoder and decoder following the settings of variational autoencoder, respectively. An evidence lower-bound (ELBO) loss is induced to optimize the model as follow

\begin{equation}
    \label{eq:elbo_tvae_loss_func}
    \log p_\theta(\mathbf{r}_j) \geq \mathop{\mathbb{E}_{q_\phi(\mathbf{z}_j | \mathbf{r}_j)}}\left[ \log p_\theta(\mathbf{r}_j | \mathbf{z}_j)\right] - \mathbb{KL} \left[ q_\phi (\mathbf{z}_j | \mathbf{r}_j) || p(\mathbf{z}_j) \right]
\end{equation}

Instead of directly calculating numerical value $\hat{c}_{i,j}$ in synthetic data $\hat{\mathbf{r}}_j$ generated from $p_\theta(\mathbf{r}_j | \mathbf{z}_j)$, the authors proposed to sample it from an intermediate distribution, $\hat{c}_{i,j} \sim \mathcal{N}(\bar\alpha_{i,j}, \delta_j)$, where $\delta_j$ is the additional learnable parameter representing the standard deviation of a numerical column $C_j$. As a result, the likelihood function is formally described as follow, with $\mathbb{CE}$ denotes the cross-entropy loss function

\begin{equation}
    \log p_\theta(\mathbf{r}_j | \mathbf{z}_j) = 
    % distribution of alpha (numerical col)
    \sum_{i}^{N_c} \log \frac{1}{\sqrt{2 \pi \sigma}} \exp{\frac{\alpha_{i,j} - \bar\alpha_{i,j}}{2 \pi \sigma^2}}
    % beta of numerical col
    + \sum_{i}^{N_c} \mathbb{CE} \left( \hat{\pmb\beta}_{i,j}, \pmb\beta_{i,j} \right)
    % d of categorical col
    + \sum_{i}^{N_d} \mathbb{CE} \left( \hat{\mathbf{d}}_{i,j}, \mathbf{d}_{i,j} \right)
\end{equation}

\subsubsection{Architecture}

The encoder architecture is

\[~\left\{\begin{array}{@{}l@{}l@{}}
 \textbf{h}_0 = \text{ReLU}(\text{FC}_{|\mathbf{r}_j| \rightarrow L_0}(\mathbf{r}_j))\\[1em]
 \textbf{h}_1 = \text{ReLU}(\text{FC}_{|\mathbf{h}_0| \rightarrow L_1}(\mathbf{h}_0))\\[1em]
 \textbf{h}_{l+1} = \text{ReLU}(\text{FC}_{|\mathbf{h}_l| \rightarrow L_{l+1}}(\mathbf{h}_l)) & \qquad 1 \leq l \leq L \\[1em]
 % \textbf{h}_2 = \text{ReLU}(\text{FC}_{256 \rightarrow 256}(\mathbf{h}_1))\\[1em]
 % \textbf{h}_3 = \text{ReLU}(\text{FC}_{256 \rightarrow 128}(\mathbf{h}_2))\\[1em]
 \mathbf{\mu} = \text{FC}_{|\mathbf{h}_L| \rightarrow L}(\mathbf{h}_L) \\[1em]
 \mathbf{\sigma} = \exp(\frac{1}{2}\text{FC}_{|\mathbf{h}_L| \rightarrow L}(\mathbf{h}_L)) \\[1em]
 q_\phi(z_j | \mathbf{r}_j) \sim \mathcal{N}(\mu, \sigma \mathbf{I})
\end{array}\right.\]

% \[~\left\{\begin{array}{@{}l@{}l@{}}
%  \textbf{h}_0 = \text{ReLU}(\text{FC}_{|\mathbf{r}_j| \rightarrow 512}(\mathbf{r}_j))\\[1em]
%  \textbf{h}_1 = \text{ReLU}(\text{FC}_{512 \rightarrow 256}(\mathbf{h}_0))\\[1em]
%  \textbf{h}_2 = \text{ReLU}(\text{FC}_{256 \rightarrow 256}(\mathbf{h}_1))\\[1em]
%  \textbf{h}_3 = \text{ReLU}(\text{FC}_{256 \rightarrow 128}(\mathbf{h}_2))\\[1em]
%  \mathbf{\mu} = \text{FC}_{128 \rightarrow 128}(\mathbf{h}_3) \\[1em]
%  \mathbf{\sigma} = \exp(\frac{1}{2}\text{FC}_{128 \rightarrow 128}(\mathbf{h}_3)) \\[1em]
%  q_\phi(z_j | \mathbf{r}_j) \sim \mathcal{N}(\mu, \sigma \mathbf{I})
% \end{array}\right.\]

The decoder architecture is
\[~\left\{\begin{array}{@{}l@{}l@{}}
 \textbf{h}_0 = \text{ReLU}(\text{FC}_{L_0 \rightarrow L_1}(\mathbf{z}_j))\\[1em]
 
 \textbf{h}_{l} = \text{ReLU}(\text{FC}_{|\mathbf{h}_{l}| \rightarrow L_{l+1}}(\mathbf{h}_{l-1})) & 1 \leq l \leq L \\[1em]

 \bar\alpha_{i,j} = \tanh(FC_{|\mathbf{h}_L| \rightarrow 1}  (\mathbf{h}_L))\\[1em]

 \hat\alpha_{i,j} \sim \mathcal{N}(\bar\alpha_{i,j}, \delta_i) & 1 \leq i \leq N_c\\[1em]
 \hat{\pmb\beta}_{i,j} \sim \text{softmax}(\text{FC}_{128 \rightarrow m_i}(\mathbf{h}_2)) & 1 \leq i \leq N_c\\[1em]
 \hat{\mathbf{d}}_{i,j} \sim \text{softmax}(\text{FC}_{128 \rightarrow |D_i|}(\mathbf{h}_2)) & 1 \leq i \leq N_d\\[1em]
 p_{\theta}(\mathbf{r}_j | z_j) = \prod_{i=1}^{N_c} \mathbb{P}(\hat\alpha_{i,j} = \alpha_{i,j}) \prod_{i=1}^{N_c} \mathbb{P}(\hat{\pmb\beta}_{i,j} = \pmb\beta_{i,j}) \prod_{i=1}^{N_d} \mathbb{P}(\hat\alpha_{i,j} = \alpha_{i,j})
\end{array}\right.\]

\subsection{Shared Tabular Variational Autoencoder (STVAE).} We observe that column standard deviation $\delta_i$ in TVAE prevents the transferability across datasets during the pretraining since each dataset deserves a corresponding set of standard deviation for its numerical columns. To address the problem, we omit this parameter and directly optimize generated numerical value $\hat{\alpha}_{i,j}$ in the likelihood function using mean squared error. Hence, we refer this model as Shared TVAE (STVAE), and the modified version of $\log p_\theta(\mathbf{r}_j | \mathbf{z}_j)$ is defined as

\begin{equation}
    \log p_\theta(\mathbf{r}_j | \mathbf{z}_j) = 
    % distribution of alpha (numerical col)
    \sum_{i}^{N_c} \mathbb{MSE} \left( \hat{\alpha}_{i,j} , \alpha_{i,j} \right)
    % beta of numerical col
    + \sum_{i}^{N_c} \mathbb{CE} \left( \hat{\pmb\beta}_{i,j}, \pmb\beta_{i,j} \right)
    % d of categorical col
    + \sum_{i}^{N_d} \mathbb{CE} \left( \hat{\mathbf{d}}_{i,j}, \mathbf{d}_{i,j} \right)
\end{equation}

\subsubsection{Architecture}
The decoder architecture is

\[~\left\{\begin{array}{@{}l@{}l@{}}
 \textbf{h}_0 = \text{ReLU}(\text{FC}_{L_0 \rightarrow L_1}(\mathbf{z}_j))\\[1em]
 
 \textbf{h}_{l} = \text{ReLU}(\text{FC}_{|\mathbf{h}_{l}| \rightarrow L_{l+1}}(\mathbf{h}_{l-1})) & 1 \leq l \leq L \\[1em]

 \hat\alpha_{i,j} = \tanh(\text{FC}_{|\mathbf{h}_L| \rightarrow 1}  (\mathbf{h}_L))\\[1em]

 \hat{\pmb\beta}_{i,j} \sim \text{softmax}(\text{FC}_{128 \rightarrow m_i}(\mathbf{h}_2)) & 1 \leq i \leq N_c\\[1em]
 \hat{\mathbf{d}}_{i,j} \sim \text{softmax}(\text{FC}_{128 \rightarrow |D_i|}(\mathbf{h}_2)) & 1 \leq i \leq N_d\\[1em]
 p_{\theta}(\mathbf{r}_j | z_j) = \prod_{i=1}^{N_c} \mathbb{P}(\hat\alpha_{i,j} = \alpha_{i,j}) \prod_{i=1}^{N_c} \mathbb{P}(\hat{\pmb\beta}_{i,j} = \pmb\beta_{i,j}) \prod_{i=1}^{N_d} \mathbb{P}(\hat\alpha_{i,j} = \alpha_{i,j})
\end{array}\right.\]

\subsection{Shared Tabular Variational Autoencoder with Metadata (STVAEM).}
To further study the transferability, we also supplement signature information to tables. Note that these information is identical for a dataset. In this work, we obtain embeddings of column names extracted from a pretrained language model as signature information, and it is concateFnated along with input data. Formally, let $\mathbf{s}_i$ denotes the signature embedding of column $i$ of dataset $\mathcal{D}$, and is concatenated to each row $\mathbf{r}_j$, thus

\begin{equation}
    \mathbf{r}_j = 
    \alpha_{0,j} \oplus \pmb\beta_{0,j} \oplus \dots 
    \oplus \alpha_{N_c, j} \pmb\beta_{N_c, j} 
    \oplus \mathbf{d}_{0,j} \oplus \dots \oplus \mathbf{d}_{N_d, j} \oplus \mathbf{s}_{0,j} \oplus \mathbf{s}_{N_c,j}
    \oplus \dots \oplus \mathbf{s}_{N_d,j}
\end{equation}

\subsection{Generation of Realistic Tabular data (GReaT).}
Transformed-based models represent tokenized data in the manner of an auto-regressive \cite{Jelinek1980InterpolatedEO, NIPS2000_728f206c}, which probability of a current token $w_k \in \mathcal{W}$ is predicted by previous observed tokens

\begin{equation}
    p(\mathbf{r}) = \prod_{k=1}^{T} p(w_k | w_1, \dots, w_{k-1})
\end{equation}

To this end, the target optimization is to maximize the probability of predicting the next token given previous tokens. Hence, any representives of generative language models can be leveraged to train. With advantages of generative language models pretrained on large corpus, GReaT \cite{borisov2023language} initializes the training from those models with the expectation to capture the contextual representation, especially when features and data are combined together, can enhance the generation of tabular data. In this work, we employ generative transfomer-decoder LLM architectures \cite{radford2018improving, radford2019language, brown2020language} represented as the distilled version of GPT as the baseline model.

\section{Data Transformation}

We denote a table (or dataset) as $\mathcal{D}$. A table contains $N_c$ continuous (numerical) columns $\{D_1, D_2, \dots, D_{N_d}\}$ and $N_d$ discrete (categorical) columns $\{D_1, D_2, \dots, D_{N_d}\}$. A row $\mathbf{r}$, which is considered as transformed data from dataset $\mathcal{D}$, follows the joint distribution, $\mathbf{r} \sim \mathbb{P}(C_{1:N_c}, D_{1:N_d})$.\\

\subsection{TVAE and CTGAN}

\textbf{Data normalization}: for CTGAN and TVAE methods \cite{xu2019modeling}, a continuous column $C_i$ is learned by a Gaussian mixture model \cite{blei2006variational} with user-defined $K$ modes as $\mathbb{P}_{C_i}(c_{i,j}) = \sum_{k=1}^{K} \mu_k \mathcal{N}(c_{i,j}; \eta_k, \phi_k)$ where $\eta_k$ denotes mode $k$ with mean $\mu_k$ and standard deviation $\phi_k$. For each mode, we calculate the density probability 

\begin{equation}
    \rho_k = \mu_k \mathcal{N}(c_{i,j}; \eta_k, \phi_k)   
\end{equation}

Then, given a value $c_{i,j}$ in $C_i$, a mode is sampled among $K$, i.e. $k^*$ to calculate to normalize the value. The chosen mode is also encoded as one-hot vector, along with the normalized value to represent the value, i.e. $c_{i,j} = \{ \alpha_{i,j}, \pmb{\beta}_{i,j}\}$. Formally,
\begin{equation}
    \alpha_{i,j} = \frac{c_{i,j} - \eta_k}{4 \phi_k}   
\end{equation}

\begin{equation}
    \pmb{\beta}_{i,j}^{(k)} = 
    \begin{cases}
      1, & \text{if}\ k = k^* \\
      0, & \text{otherwise}
    \end{cases}
\end{equation}

For a categorical value $d_{i,j}$ in $D_i$, it is represented by a one-hot vector $\mathbf{d}_{i,j}$

\begin{equation}
    \mathbf{d}_{i,j}^{(l)} = 
    \begin{cases}
        1, & \text{if}\ l = l^* \\
        0, & \text{otherwise}
    \end{cases}
\end{equation}

As a result, after the transformation, a row is represented as the concatenation of transformed data

\begin{equation}
    \mathbf{r}_j = \alpha_{1,j} \oplus \pmb{\beta}_{1,j} \oplus \dots \oplus \alpha_{N_c, j} \pmb{\beta}_{N_c, j} \oplus \mathbf{d}_{1,j} \oplus \dots \oplus \mathbf{d}_{N_d, j}
\end{equation}

\subsection{Transformers}
Intuitively, transformer-based models such as GReaT \cite{borisov2023language} expect tabular data to be converted into sequence of words. In consideration of that, a textual encoding paradigm is often leveraged. Taking the transformation of GReaT into account, the feature (or table column) name $f_i$ and data of column $D_i$, $C_i$ is employed to construct to desbribe as a sentence. For a value at row $j$ belonging to a column $i$ of $D_i$ or $C_i$, a textual representation $t$ is constructed as follow

\begin{equation}
    t_{j,i} = f_i \text{ "is" } D_{j,i}
\end{equation}

\begin{equation}
    \mathbf{t}_{j} = t_0 \oplus t_1 \oplus \dots \oplus t_{N_d} \oplus \dots \oplus t_{N_c}
\end{equation}

Since tabular data features are order-independent, an order feature permutation is applied to randomly shuffle $\mathbf{t}_{j}$. Afterward, transformed data is represented as tokens by a vocabulary $\mathcal{W}$

\begin{equation}
    \mathbf{r}_j = \text{TOKENIZE}(\mathbf{t}_j)
\end{equation}

Where $\text{TOKENIZE}$ denotes any tokenization methods such as Byte-Pair-Encodings \cite{sennrich-etal-2016-neural}.

\section{Data Creation}
\subsection{Data Acquisition}
\subsubsection{Kaggle Dataset}
 Kaggle is a prominent platform for data science enthusiasts and professionals, hosting huge datasets spanning diverse domains, including healthcare, finance, and more. The process of acquiring these tabular datasets desires a systematic approach, wherein metadata descriptions, dataset versions, and associated documentation are carefully parsed and analyzed. This ensures the integrity and comprehensiveness of the acquired data, laying a robust foundation for subsequent data cleaning and preprocessing. The process of crawling and cleaning Kaggle datasets is described below. 
 
 %Leveraging Kaggle's extensive repository facilitates access to high-quality, curated datasets and fosters collaborative research by enabling the sharing and dissemination of valuable data resources across the scientific community. Harnessing tabular datasets from Kaggle is invaluable in advancing data-driven research across various disciplines. 

Step 1: Pre-filtering Kaggle Tabular Datasets: to streamline the dataset acquisition process, we first filter the Kaggle datasets based on file type, only datasets with the file type CSV are considered for further processing. Subsequently, usability rating is considered, specifically datasets with a minimum usability rating of 8.00 or higher are prioritized, ensuring the quality and reliability of the selected datasets.

Step 2: Automated Crawling Dataset URLs. We utilize Selenium, a powerful web scraping tool, to automate the process of extracting dataset links from Kaggle. This automated crawling ensures efficient retrieval of dataset URLs, reducing manual effort and potential errors in the dataset acquisition process.

Step 3: Downloading Datasets Using Kaggle API.
After extracting the dataset links, we employ the Kaggle API to download the selected datasets pro-grammatically. This automated download process facilitates batch downloading of datasets, further enhancing the efficiency of the acquisition pipeline. After completing steps 1-3, we found 43514 potential datasets. The following data cleaning is discussed in the next section which resulting in 1435 high quality tables with only numerical and categorical values with the statistics summarised in Table \ref{tab:statistics}.

\begin{figure}[h]
    \centering
    \includegraphics[width=1.0\textwidth]{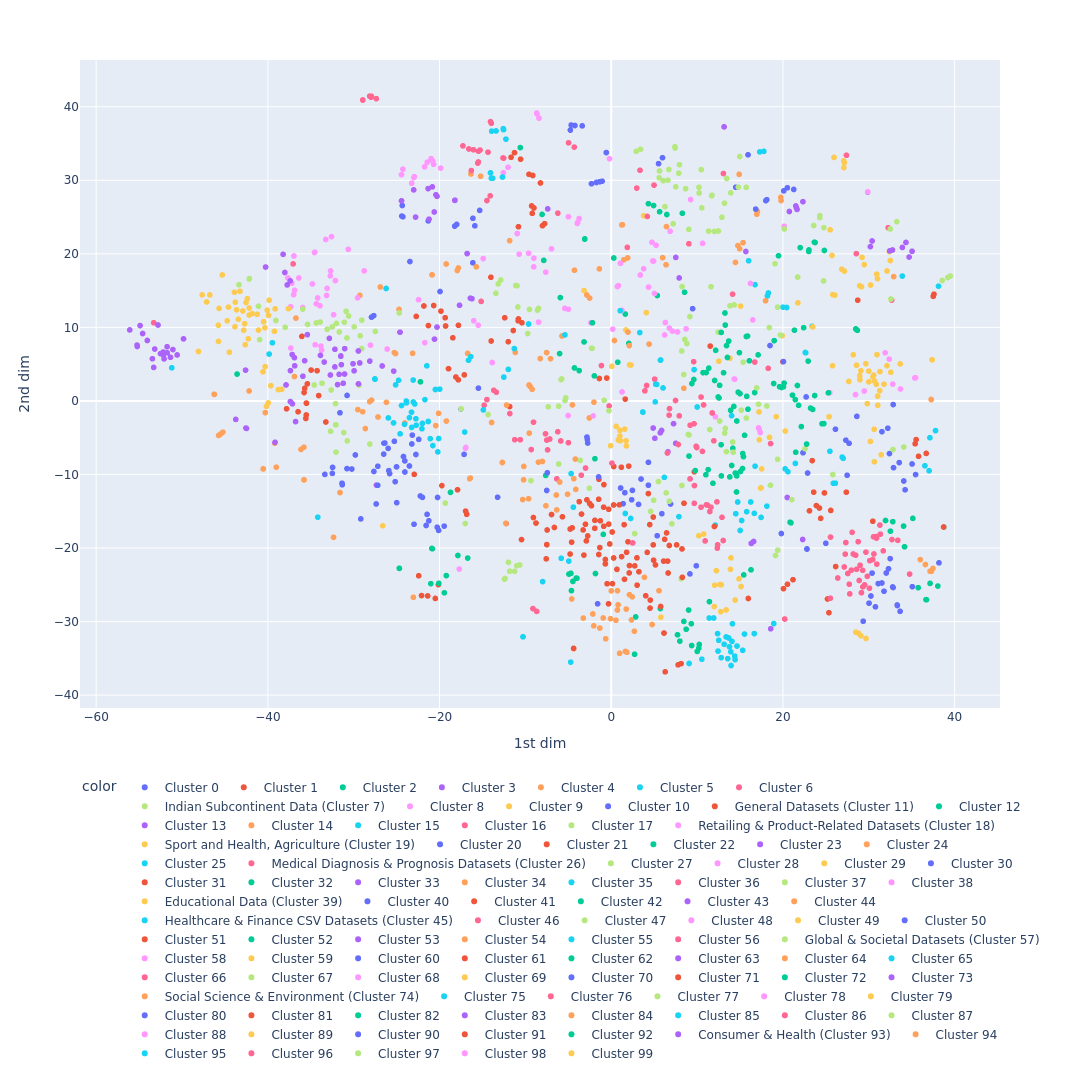}
    \caption{TSNE representation of total 100 domains clustered by K-Means algorithm, including manually labeled clusters.}
    \label{fig:domain kmean top 100}
\end{figure}

We utilized the $k$-means algorithm to cluster BERT embeddings of table names, setting $K$ to 100. Figure \ref{fig:domain kmean top 10} showcases the top 10 clusters containing the most elements. The clustering effectively categorized the data by table names, with each cluster representing specific domains such as consumer, health, and education data. However, as shown in Figure \ref{fig:domain kmean top 100}, distinguishing domains purely based on table names within 100 clusters presents a challenge that requires future work. 

\begin{table}[]
\centering
\caption{\label{tab:statistics}Statistics about the datasets}
\begin{tabular}{llllll}
\hline
\textbf{Dataset} & \textbf{Raw tables} & \textbf{\begin{tabular}[c]{@{}l@{}}Cleaned tables \\ train/val/test\end{tabular}} & \textbf{Avg. \# columns} & \textbf{Avg. \# rows}  & \textbf{License}\\ \hline
Kaggle Random &    43514   & 1148/143/144      & 8.37      & 224.8   &  Kaggle         \\ \hline
Kaggle Domains &    43514   & 969/218/248      & 8.37      & 224.8   & Kaggle           \\ \hline
GitTables &   1M  & 1006/126/126      & 9.51      & 1112.68   & CC BY 4.0          \\ \hline
\end{tabular}
\end{table}

\subsubsection{GitTables Dataset}

GitTables \cite{hulsebos2023gittables} is a large-scale corpus of relational tables extracted from CSV files on GitHub, designed to facilitate the development of table representation models and applications in areas such as data management and data analysis. 
The corpus includes 1 million raw tables. Despite their tabular format, the majority of tables in the GitTables dataset contain unstructured data such as text (e.g., product reviews) or time series (e.g. server bandwidth). This presents a challenge in building foundational models for tabular data, requiring a significant effort in data cleaning. After removing tables with sole structured information there are 1258 tables split into train/val/test as illustrated in Table \ref{tab:statistics}.

\subsection{Data Cleaning}
We recognize that the tabular datasets from both Kaggle and GitTables are quite noisy. Although they are in a structured format, most of the tables include a mix of structured information such as text, temporal and spatial data, along with irrelevant values like ID columns, missing values, and formatting errors. Below, we summarize the main steps of data cleaning. Some of these steps are automated, while a few require manual inspection and curation.

\textbf{Files merging and cleaning}
After downloading the tables, we discovered variations in their structures. For examples, some datasets comprise CSV files serving as annotation labels for tasks like computer vision or natural language processing, such as image classification, with two columns: image path and label, or question answering in NLP, with two columns: text and IOB-label. Conversely, other datasets consisted of multiple tables, necessitating manual merging to create a comprehensive final table. This manual processing demands a profound understanding of the data and its associated task. To ensure suitability for pre-training the TabularFM task, we conducted a data cleaning and validation pipeline:
\begin{itemize}
    \item Single Tabular File: We check if the downloaded dataset contains only one tabular file of CSV type. If the dataset fails this criterion, it is masked for manually merging. 
    \item Manually merge: In cases the dataset fails the Single Tabular File criterion, we check that which data has a clear description about the dataset (have clear metadata about columns name, discription, columns type,) in Kaggle platform. We manually merged the columns and saved the final table to a single CSV file. 
    \item Automated Validation of Annotation CSV Tables: As mentioned above, datasets containing CSV file annotation labels for computer vision or NLP downstream tasks are unsuitable for pre-training TabularFM. However, these datasets consistently have discernible patterns, typically characterized by all columns only contains two types of column: categories (text type labels), or texts (long string, paths, URLs). Leveraging this consistency, we implement automated validation procedures to effectively filter datasets having these characteristics.
\end{itemize}

\textbf{Data Filtering Based on Column Types} To optimize the quality of the acquired datasets, we apply a filtering step based on the column types. We remove columns containing non-numeric and non-categorical data types, including long string types, path strings, links, URLs, phone numbers, etc.

\textbf{Noisy column identification, missing value imputation} To ensure the effective training, we further construct automated preprocessing across datasets. For each column in a dataset, we automatically process the data based on the following criteria

\begin{itemize}
    \item Remove identity data: since identity has no meaning in generation process, we discard columns with identity values such as id columns.
    
    \item Timestamp columns: any data representing date or timestamp is also removed. Since timestamp data is specialized for time-series generative models, which is out of scope for this work.
    
    \item Remove sparse categorical columns: we calculate frequency of each category. We discard the column if the number of categories is significantly large, i.e. greater than 90\%, compared to the total number of samples. Furthermore, if the average frequency across categories is less than a threshold, we also discard the column. For the two proposed datasets, we set this threshold as 3\%.
    
    \item Data imputation: empty cells in tables are imputed with conventional strategy. Particularly, if the total number of null values exceed a threshold, which is set to 50\% in this work, the column will be discarded. Otherwise, we impute the data depends on the data type. For numerical column, average data is calculated, for categorical column, we impute the value of highest mode category.
\end{itemize}

Afterwards, a dataset may lose large number of columns. We then further discard a table if there are few columns left compared to a threshold. We eliminate tables which number of unqualified columns exceeds 90\% in this work.

\section{Training Details}

Let $\pmb{\mathcal{D}} = \{\mathcal{D}_i\}$ be the whole datasets across domains. We divided them into three kinds of datasets as $\pmb{\mathcal{D}}^{\text{pretrain}}$, $\pmb{\mathcal{D}}^\text{val}$, $\pmb{\mathcal{D}}^\text{test}$ correpsonding to pretraining datasets, validation and test datasets, respectively. Our objective is to study the transferability of tabular datasets. To this end, we train a model on $\pmb{\mathcal{D}}^{\text{pretrain}}$ to achieve a pretrained model, the model is then finetuned on each dataset of $\pmb{\mathcal{D}}^\text{val}$ and $\pmb{\mathcal{D}}^\text{test}$. Along with that, we initialize a corresponding model for each dataset and train from scratch, i.e. single training. We then evaluate and compare the performance of fine-tuning versus training from scratch.

\subsection{Pretraining}
Let $\pmb{\mathcal{M}}^p$ be the model training on $\pmb{\mathcal{D}}^\text{pretrain}$. To avoid catastrophic forgetting \cite{}, for each iteration, we train the model across dataset $\pmb{\mathcal{D}}^\text{pretrain}$ with a single epoch and repeat for the next iteration. We randomly shuffle $\pmb{\mathcal{D}}^\text{pretrain}$ for every iteration. The pretraining procedure can be describe in detail as follow

\begin{enumerate}
    \item Initialize $\pmb{\mathcal{M}}^p$ 
    \item For each \texttt{iteration}
    \begin{enumerate}
        \item Randomly shuffle $\pmb{\mathcal{D}}^\text{pretrain}$
        \item For each dataset $\mathcal{D}_i \in \pmb{\mathcal{D}}^\text{pretrain}$, transform and fit data to $\pmb{\mathcal{M}}^p$ with \texttt{epoch = 1}
    \end{enumerate}
\end{enumerate}

\subsection{Fine-tuning and Single Training}

For each dataset $\mathcal{D}_i \in \pmb{\mathcal{D}}^\text{val / test}$, we initialize $\pmb{\mathcal{M}}_i^\text{ft}$ with weights from $\pmb{\mathcal{M}}^p$ and finetune the model. Our framework employs early stopping on loss of validation set of each dataset to prevent overfitting. For GAN-based models, since it is challenging to early stop the training only observing the validation loss, we keep checkpoints during the training.For single training, we initialize a model $\pmb{\mathcal{M}}_i^\text{st}$ and train from scratch for each $\mathcal{D}_i \in \pmb{\mathcal{D}}^\text{val / test}$. We also leverage early stopping in the training except for GAN-based models. \\

In summary, for each dataset $\mathcal{D}_i \in \pmb{\mathcal{D}}^\text{val / test}$, the training procedure of finetuning is described as follow 

\begin{enumerate}
    \item Initialize $\pmb{\mathcal{M}}_i^\text{ft}$ weights from $\pmb{\mathcal{M}}^p$
    \item For each \texttt{epoch}
    \begin{enumerate}
        \item Transform and fit data of $\mathcal{D}_i$ to $\pmb{\mathcal{M}}^\text{ft}$
        \item Early stopping checking
        \item Checkpoint checking
    \end{enumerate}
\end{enumerate}

For each dataset $\mathcal{D}_i \in \pmb{\mathcal{D}}^\text{val / test}$, the training procedure of single training is
\begin{enumerate}
    \item Initialize $\pmb{\mathcal{M}}_i^\text{st}$ from scratch 
    \item For each \texttt{epoch}
    \begin{enumerate}
        \item Transform and fit data of $\mathcal{D}_i$ to $\pmb{\mathcal{M}}^\text{st}$
        \item Early stopping checking
        \item Checkpoint checking
    \end{enumerate}
\end{enumerate}
\subsection{Evaluation}
To evaluate the performance of tabular generation, we compare the quality of synthetic data versus real data by two properties. We measure the distribution of columns data between synthetic and real data, called \textit{Column Shapes}. Correlation among columns is also computed, then compare the difference between ones of synthetic and real data, called \textit{Column Trends}. We then average the two metrics as \textit{Overall Score}. Formally, given a dataset $\mathcal{D} = \{ D_0, D_1, \dots, D_{N_d}, C_0, C_1, \dots, C_{N_c} \}$, the overall evaluation score, and score of column shape, trend are calculated as

\begin{equation}
    \mathcal{S}^{(\mathcal{D})} = \frac{\mathcal{S}_\text{shape}^{(\mathcal{D})} + \mathcal{S}_\text{trend}^{(\mathcal{D})}}{2}
\end{equation}

\begin{equation}
    \mathcal{S}_\text{shape}^{(\mathcal{D})} = \frac{1}{N_c + N_d} \sum_{c \in N_c, N_d} \mathcal{S}_\text{shape}^{(c)}
\end{equation}

\begin{equation}
    \mathcal{S}_\text{trend}^{(\mathcal{D})} = \frac{1}{N_c + N_d} \sum_{c \in N_c, N_d} \mathcal{S}_\text{trend}^{(c)}
\end{equation}

We describe the definition of scores to calculate on the table columns as follows. To avoid complication, we omit the annotation of dataset and columns, \textit{syn} and \textit{real} are added to represent the synthetic data and real data. In practice, the following metrics are computed separately for finetuned model and trained from scratch model. We then evaluate the performance between model finetuned from a pretraining, and model trained from scratch by comparing between those measurements.
% In practice, $\text{syn}$ is calculated for synthetic data of finetuned model, and trained from scratch model separately.

\subsubsection{Column Shapes}

For numerical data, we leverage a non-parametric statistical method, which is Kolmogorov-Smirnov test \cite{massey1951kolmogorov}, to represent the fitness of shapes between distribution of synthetic and real data. 
Formally, let $F_\text{syn}(x)$ and $F_\text{real}(x)$, $x \in C_i$, denote the Cumulative Distribution Functions over distribution data of numerical columns of synthetic and real data, respectively. The score $\mathcal{S}_\text{shape} \in [0, 1]$ represents the similarity between distribution of numerical synthetic and real data is described as follow (higher is better)

\begin{equation}
    \mathcal{S}_\text{shape} = 1 - \sup_{x} |F_\text{syn}(x) - F_\text{real}(x)|
\end{equation}

For categorical data, Total Variance Distance is computed to measure the column shapes. To this end, let $\Omega$ be the space of categories of $D_i$, we calculate the ratio of each category $w \in \Omega$, then measure the difference between ones of synthetic and real data. As a result, the score of column shape is adjusted to express higher is better as follow

\begin{equation}
    \mathcal{S}_\text{shape} = 1 - \frac{1}{2} \sum_{w \in \Omega} |R_\text{syn}(w) - R_\text{real}(w)|
\end{equation}

\subsubsection{Column Trends}
To measure the closeness of column pair trends between real and synthetic data, we measure every column pairs first and calculate the average. To this end, correlation is computed among columns in real and synthetic data separately, and then measure the difference. Depend on the data types, different methods are applied.\\

To measure the correlation between two numerical column, we leverage Pearson correlation \cite{freedman2007statistics}. Let $\rho(C_m, C_n)$ denote the Pearson correlation between two numerical columns $C_m, C_n \in D$,

\begin{equation}
    \rho(C_m, C_n) = \frac{\mathbb{E} \left( (C_m - \mu_{C_n}) (C_n - \mu_{C_n}) \right)}{\sigma_{C_m} \sigma_{C_n}}
\end{equation}

where $\mu_*$ and $\sigma_*$ represents the mean and standard deviation of column. As a result, $\rho_\text{syn}(C_m, C_n)$ and $\rho_\text{real}(C_m, C_n)$ is obtained, the column trend score between two numerical columns is defined as

\begin{equation}
    \mathcal{S}_\text{trend} =1 - \frac{|\rho_\text{syn}(C_m, C_n) - \rho_\text{real}(C_m, C_n)|}{2}
\end{equation}

In the case of two categorical columns, we calculate a normalized contingency table. In other words, we compute the number of samples of any category pairs $w_m, w_n \in \Omega$, then normalize by total number of samples, each combination is represented by $R(w_m, w_n)$. After that, Total Variance Distance is applied to calculate the column trend score. Given $D_m, D_n \in D$ as pair of categorical columns. The correlation (higer is better) is express as

\begin{equation}
\label{eq:s_trend_cat}
    \mathcal{S}_\text{trend} = 1 - \frac{1}{2} \sum_{w_m \in \Omega_{D_m}} \sum_{w_n \in \Omega_{D_n}} |R_\text{syn}(w_m, w_n) - R_\text{real}(w_m, w_n)|
\end{equation}
When dealing with column pairs of categorical and numerical data types, we first transform numerical to categorical data by grouping it into bins, and calculate the column trend score by \eqref{eq:s_trend_cat}.

\end{document}